\documentclass{llncs}
\usepackage{graphicx}
\usepackage{amsmath,amssymb} 
\usepackage{color}

\usepackage{graphicx}
\usepackage{amsmath}
\usepackage{amssymb}
\usepackage{booktabs}

\usepackage{color, colortbl}
\usepackage{algorithm}
\usepackage{algorithmic}
\usepackage{amsmath}
\usepackage{multirow}
\usepackage{xcolor}
\usepackage{xr-hyper}
\usepackage{cite}
\usepackage{listings}
\usepackage{xspace}

\usepackage[pagebackref,breaklinks,colorlinks]{hyperref}

\usepackage[capitalize]{cleveref}
\crefname{section}{Sec.}{Secs.}
\Crefname{section}{Section}{Sections}
\Crefname{table}{Table}{Tables}
\crefname{table}{Tab.}{Tabs.}
\makeatletter
\DeclareRobustCommand\onedot{\futurelet\@let@token\@onedot}
\def\@onedot{\ifx\@let@token.\else.\null\fi\xspace}

\def\eg{\emph{e.g}\onedot} 
\def\ie{\emph{i.e}\onedot}

\def\etal{\emph{et al}\onedot}
\makeatother
\newcommand{\rowgray}{\rowcolor{gray!7}}
\begin{document}
\pagestyle{headings}
\mainmatter

\def\ACCV22SubNumber{29}  

\title{Revisiting Image Pyramid Structure for High Resolution Salient Object Detection} 
\titlerunning{ACCV-22 submission ID \ACCV22SubNumber}
\authorrunning{ACCV-22 submission ID \ACCV22SubNumber}

\author{Taehun Kim, Kunhee Kim, Joonyeong Lee, Dongmin Cha, Daijin Kim}
\institute{Pohang University of Science and Technology}

\titlerunning{InSPyReNet}
\author{Taehun Kim\inst{1} \and
Kunhee Kim\inst{1}  \and
Joonyeong Lee\inst{1} \and
Dongmin Cha\inst{1}   \and
Jiho Lee\inst{1}   \and
Daijin Kim\inst{1}
}
\authorrunning{Kim et al.}
\institute{$^1$Dept. of CSE, Pohang University of Science and Technology (POSTECH), Korea \\
\email{\{taehoon1018, kunkim, joonyeonglee, cardongmin, jiholee, dkim\}@postech.ac.kr} \\
\url{https://github.com/plemeri/InSPyReNet.git}} 

\maketitle

\begin{abstract}
  Salient object detection (SOD) has been in the spotlight recently, yet has been studied less for high-resolution (HR) images. 
  Unfortunately, HR images and their pixel-level annotations are certainly more labor-intensive and time-consuming compared to low-resolution (LR) images and annotations.
  Therefore, we propose an image pyramid-based SOD framework, Inverse Saliency Pyramid Reconstruction Network (InSPyReNet), for HR prediction without any of HR datasets.
  We design InSPyReNet to produce a strict image pyramid structure of saliency map, which enables to ensemble multiple results with pyramid-based image blending.
  For HR prediction, we design a pyramid blending method which synthesizes two different image pyramids from a pair of LR and HR scale from the same image to overcome effective receptive field (ERF) discrepancy.
  Our extensive evaluations on public LR and HR SOD benchmarks demonstrate that InSPyReNet surpasses the \textit{State-of-the-Art} (SotA) methods on various SOD metrics and boundary accuracy.
\end{abstract}
\section{Introduction}
    \label{sec:intro}

\begin{figure}
    \centering
    \includegraphics[width=\linewidth]{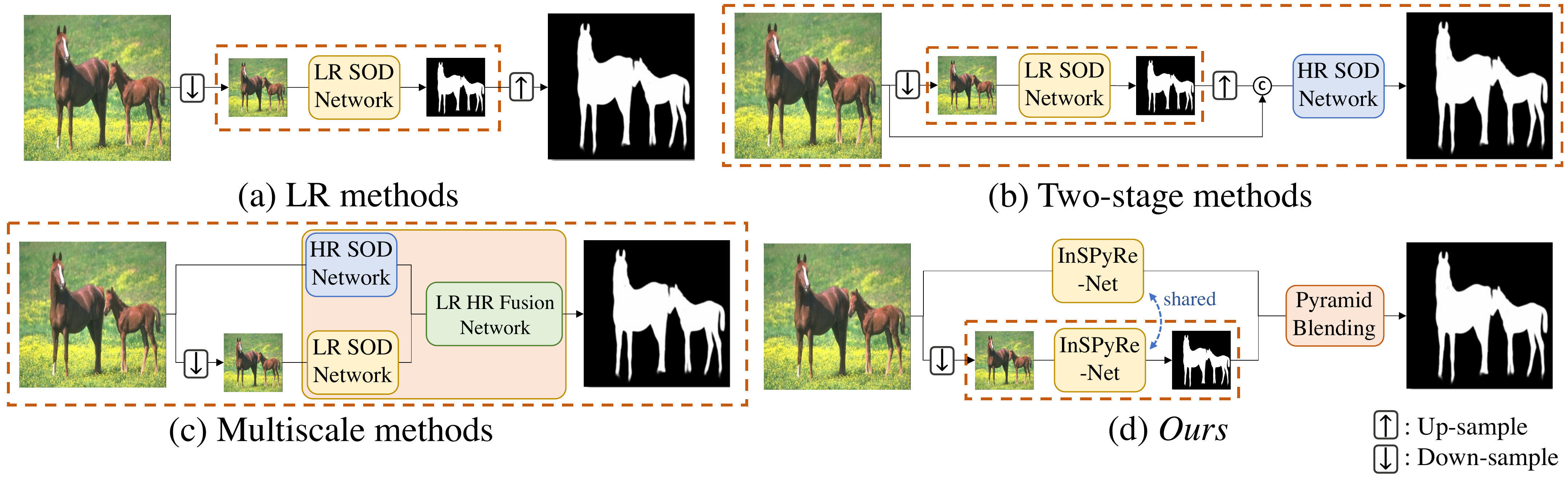} 
    \caption{Different approaches for HR SOD prediction. Areas denoted as a dashed box are trained with supervision.
    (a): Resizing HR input to LR, then up-sample. Works for any methods, lack of details. 
    (b): Requires multiple training sessions, and HR datasets~\cite{zeng2019towards,tang2021disentangled}. 
    (c): Can overcome ERF discrepancy, but the architecture is complex, requires HR datasets~\cite{xie2022pyramid}.
    (d): Works without HR dataset training. We predict multiscale results with single network and synthesize HR prediction with pyramid blending.}
    \label{fig:1}
\end{figure}

    While there are many successful works for SOD in low-resolution (LR) images, there are many demands on high-resolution (HR) images.
    One can argue that methods trained with LR datasets produce decent results on HR images by resizing the input size (\cref{fig:1}a), 
    but the quality in terms of the high-frequency details of prediction still remains poor in that way.
    Moreover, previous studies on HR prediction have been working on developing complex architectures and proposing laborious annotations on HR images~\cite{zeng2019towards, tang2021disentangled, zhang2021looking, xie2022pyramid} (\cref{fig:1}b, c). 

    In this paper, we focus on only using LR datasets for training to produce high-quality HR prediction.
    To do so, we mainly focus on the structure of saliency prediction, which enables to provide high-frequency details from the image regardless of the size of the input.
    However, there is still another problem to be solved where the effective receptive fields (ERFs)~\cite{luo2016understanding} of HR images are different from the LR images in most cases.
    To alleviate the aforementioned issues, we propose two solid solutions which are mutually connected to each other.
    
\begin{figure}
    \centering
    \includegraphics[width=\linewidth]{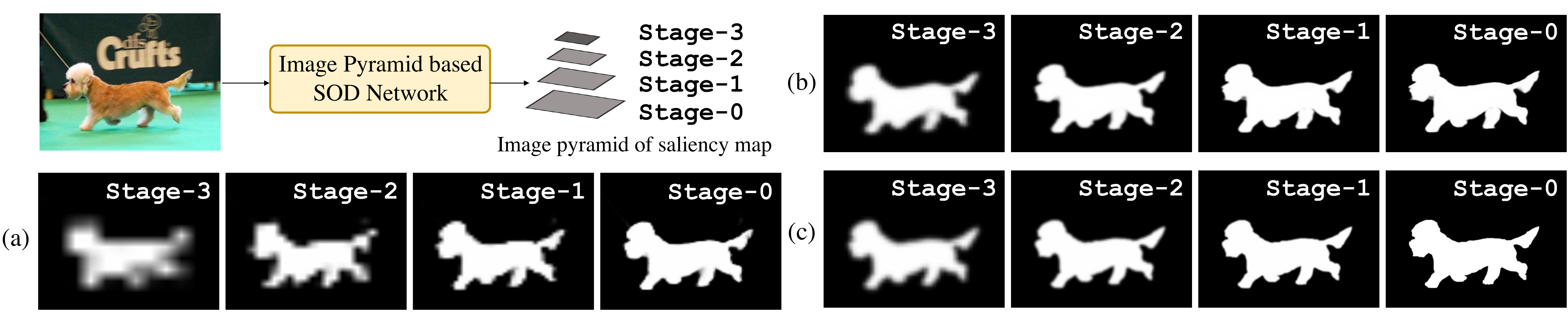} 
    \caption{Comparison of image pyramid based saliency map between (a) Chen \etal~\cite{chen2018reverse}, (b) InSPyReNet, and image pyramid of (c) ground truth.
    Compared to the image pyramid of ground truth saliency map, Chen \etal shows distorted results especially for the higher stages (\eg, \texttt{Stage-3}).
    However, our InSPyReNet shows almost identical results compared to the ground truth across each stage.}
    \label{fig:2}
\end{figure}

    First is to design a network architecture which enables to merge multiple results regardless of the size of the input.
    Therefore, we propose Inverse Saliency Pyramid Reconstruction Network (InSPyReNet), which predicts the image pyramid of the saliency map. 
    Image pyramid is a simple yet straightforward method for image blending~\cite{burt1983mosaics}, so we design InSPyReNet to produce the image pyramid of the saliency map directly.
    Previous works have already used image pyramid prediction, but results did not strictly follow the structure, and hence unable to use for the blending (\cref{fig:2}). 
    Therefore, we suggest new architecture, and new supervision techniques to ensure the image pyramid structure which enables stable image blending for HR prediction.
    
    Second, to solve the problem of ERF discrepancy between LR and HR images, we design a pyramid blending technique for the inference time to overlap two image pyramids of saliency maps from different scales.
    Recent studies of HR SOD methods use two different scales of the same image, by resizing HR image to LR, to alleviate such problem~\cite{zhang2021looking, xie2022pyramid}, but the network should be complicated and large (\cref{fig:1}c).
    Simply forwarding HR images to the InSPyReNet, or other LR SOD networks fail to predict salient region since they are not trained with HR images.
    Nevertheless, we notice the potential of enhancing details for high-quality details from HR prediction, even the result shows a lot of False Positives (HR prediction in~\cref{fig:3}).
    To combine the robust saliency prediction and details from LR and HR predictions, we blend the two image pyramids of saliency maps.

    InSPyReNet \textit{does not require HR training and datasets}, yet produces high-quality results on HR benchmarks.
    A series of quantitative and qualitative results on HR and LR SOD benchmarks show that our method shows SotA performance, 
    yet more efficient than previous HR SOD methods in terms of training resources, annotation quality, and architecture engineering.

\section{Related Works}
\label{sec:rel}

\noindent
\textbf{Salient Object Detection.} 
Edge-Based Models are studied in SOD for better understanding of the structure of the salient object by explicitly modeling the contour of the saliency map.
Methods with auxiliary edge estimator require additional edge GT, or extra training process with extra edge datasets. 
For instance, EGNet~\cite{zhao2019egnet} has an additional edge estimation branch which is supervised with additional edge-only dataset.
However, the effect of edge branch is limited to the encoder network (backbone), expecting better representation with robust edge information, 
because the estimated edge from the edge branch is not directly used to the detection. 
Also, LDF~\cite{wei2020label} designed an alternative representation for the edge information. They divided the saliency map into `body' and `detail', which corresponds to the edge part. 
Unlike EGNet, they utilized both `body' and `detail' for the saliency prediction in the inference stage. 
However, to achieve the disentanglement of `body' and `detail' components, it requires multiple training stages and ground truth generation.

Unlike auxiliary edge models, we embedded the image pyramid structure to the network for saliency prediction, 
which does not require additional training process nor extra datasets, and the decoder network is implicitly trained to predict the Laplacian of the saliency map, 
high-frequency details of the larger scales, which implicitly includes edge information. Thanks to this simple structure, we also do not require additional training stages.

\begin{figure}
    \centering
    \includegraphics[width=\linewidth]{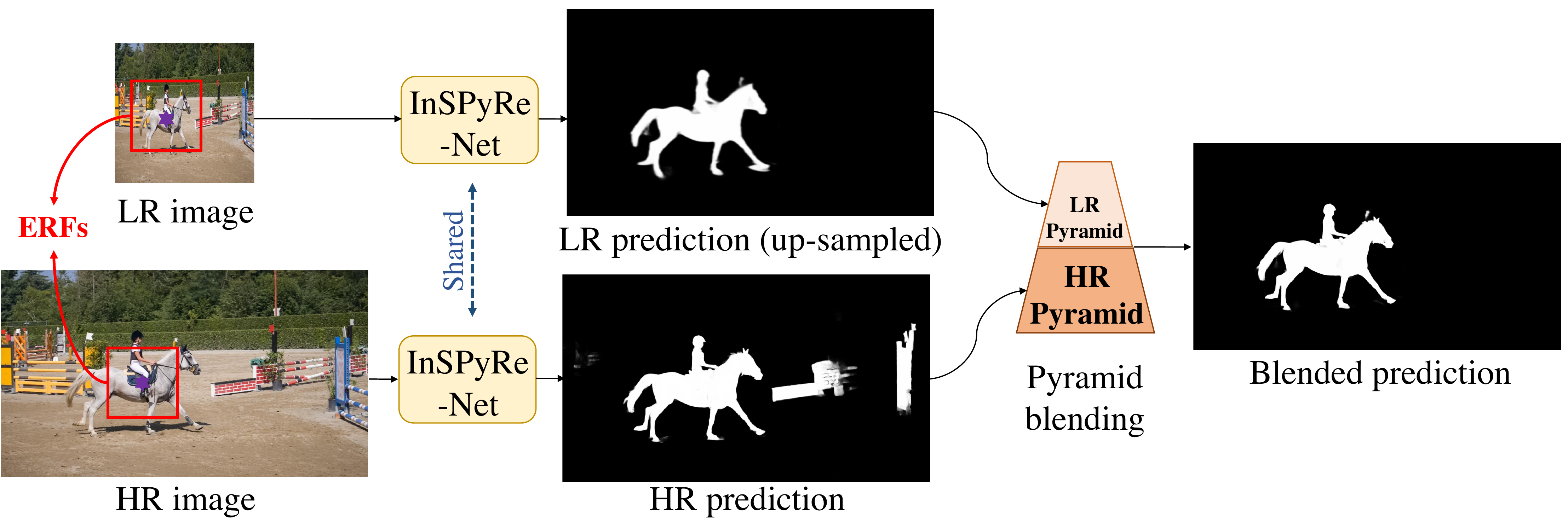} 
    \caption{Illustration of effective receptive field (ERF)~\cite{luo2016understanding} discrepancy between LR and HR images from InSPyReNet.
    LR prediction shows successful saliency prediction, but lack of details. While HR prediction shows better details but due to the ERF discrepancy (red boxes), it over detects objects.
    With pyramid blending, we can capture the global dependency from LR prediction while enhance local details from HR prediction at the same time. \textit{Best viewed by zooming in.}}
    \label{fig:3}
\end{figure}

\noindent
\textbf{Image Segmentation for HR Images.} Pixel-wise prediction tasks such as SOD resize input images into a pre-defined shape (\eg, $384\times384$) for batched and memory efficient training.
This is plausible since the average resolution of training datasets are usually around 300 to 400 for both width and height.
For example, the average resolution is $378 \times 469$ for ImageNet~\cite{russakovsky2015imagenet}, and $322\times372$ for DUTS~\cite{wang2017learning}.
After training, resizing input images into a pre-defined shape is often required, especially when the input image is relatively larger than the pre-defined shape (\cref{fig:1}a).
However, down-sampling large images causes severe information loss, particularly for high-frequency details. 
We can overcome this problem by not resizing images, but current SotA SOD methods fail to predict appropriate saliency map because they are neither trained with HR dataset nor designed to produce HR prediction.
Most likely, the problem is the discrepancy between the effective receptive fields~\cite{luo2016understanding} of the same corresponding pixel from the original and resized images (\cref{fig:3}).

CascadePSP~\cite{wu2019cascaded} first tackled this problem in semantic segmentation by approaching HR segmentation by a refinement process.
They trained their model with coarse segmentation masks as an input with a set of augmentation techniques, 
and used the model to refine an initial segmentation mask with multiple global steps and local steps in a recursive manner.
However, they need an initial prediction mask from standalone models~\cite{zhao2017pyramid,chen2017rethinking}, which is definitely not resource-frendly.
Zeng \etal~\cite{zeng2019towards} first proposed HR dataset for SOD task with a baseline model which consists of separate LR, HR and fusion networks. 
They combined global (GLFN) and local (LRN) information by two separate networks dedicated for each of them.
Tang \etal~\cite{tang2021disentangled} also designed LR and HR networks separately, where the branch for the HR (HRRN) gets an image and a predicted saliency map from the LR branch (LRSCN).
PGNet~\cite{xie2022pyramid} first proposed a standalone, end-to-end network for HR prediction by combining features from LR and HR images with multiple backbone networks.

Aforementioned methods require HR datasets for training, complex model architecture, multiple training sessions for submodules (\cref{fig:1}b, c).
Unlike previous methods, InSPyReNet does not require HR datasets for training, yet predicts fine details especially on object boundary.

\noindent
\textbf{Image Pyramid in Deep Learning Era.} Studies of pixel-level prediction tasks have shown successful application of image pyramid prediction.
LapSRN~\cite{lai2017deep} first applied a Laplacian image prediction for Super Resolution task and since then, most end-to-end supervised super resolution methods adopt their structure.
LRR~\cite{ghiasi2016laplacian} first applied a Laplacian image pyramid for the semantic segmentation task in the prediction reconstruction process.
Then, Chen \etal~\cite{chen2018reverse} adopted LRR prediction strategy for the SOD\@ with reverse attention mechanism, and UACANet~\cite{kim2021uacanet} extended self-attention mechanism with uncertainty area for the polyp segmentation.
As the above methods have already proved that without any training strategy, we can expect the network to implicitly predict the image pyramid by designing the architecture.
However, without extra regularization strategy for the supervision to follow the image pyramid structure rigorously, we cannot make sure that the Laplacian images from each stage truly contains high-frequency detail (\cref{fig:2}).

We revisit this image pyramid scheme for prediction, and improve the performance by setting optimal stage design for image pyramid and regularization methods to follow pyramidal structure.
Also, to the best of our knowledge, InSPyReNet is the \textit{first attempt to extend image pyramid prediction for multiple prediction ensembling by image blending technique.} 
This is because previous methods' Laplacian images did not strictly follow actual high-frequency detail. Rather, they focus more on correcting errors from the higher stages (\cref{fig:2}).
Unlike previous methods, we adopt scale-wise supervision (\cref{fig:4}b), Stop-Gradient and pyramidal consistently loss (\cref{sec:met.4}) for regularization which enables consistent prediction, 
and hence we are able to use blending technique by utilizing multiple results to facilitate more accurate results on HR benchmarks.
\section{Methodology}
\label{sec:met}

    \subsection{Model Architecture}
        \noindent
        \textbf{Overall Architecture.} We use Res2Net~\cite{gao2019res2net} or Swin Transformer~\cite{liu2021swin} for the backbone network, but for HR prediction, we only use Swin as a backbone.
        We provide a thorough discussion (\cref{sec:disc}) for the reason why we use only Swin Transformer for HR prediction.

        From UACANet~\cite{kim2021uacanet}, we use Parallel Axial Attention encoder (PAA-e) for the multiscale encoder to reduce the number of channels of backbone feature maps 
        and Parallel Axial Attention decoder (PAA-d) to predict an initial saliency map on the smallest stage (\ie, \texttt{Stage-3}).
        We adopt both modules because they capture global context with non-local operation, and it is efficient thanks to the axial attention mechanism~\cite{ho2019axial,wang2020axial}.
        
\begin{figure*}[t]
    \centering
    \includegraphics[width=\textwidth]{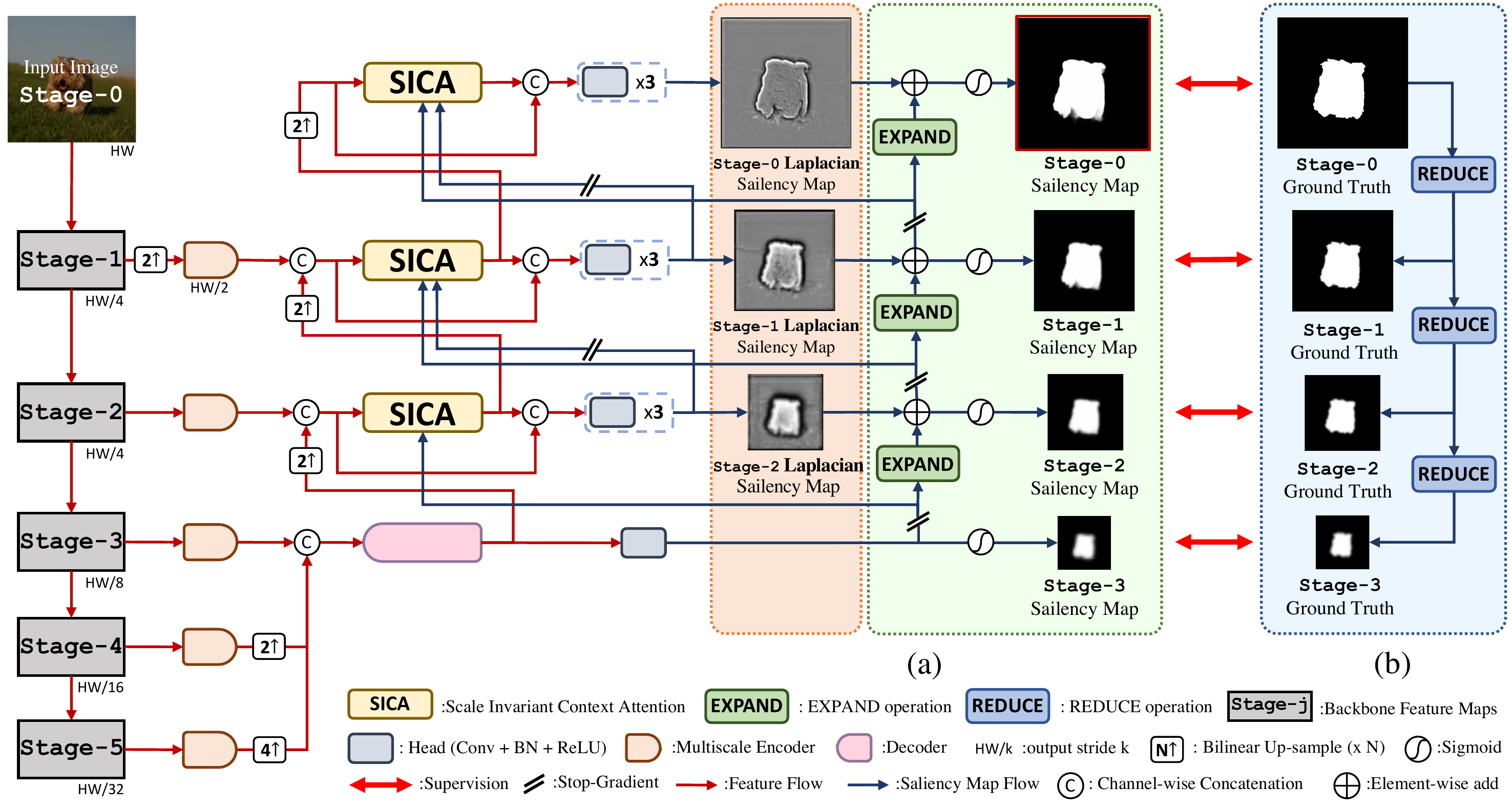} 
    \caption{The architecture of proposed InSPyReNet. 
            (a) The initial saliency map from \texttt{Stage-3} and Laplacian saliency maps from higher-stages are combined with EXPAND operation to be reconstructed to the original input size. 
            (b) The ground-truth is deconstructed to the smaller stages for predicted saliency maps from each stage by REDUCE operation.}
    \label{fig:4}
\end{figure*}

        Refer to the stage design in~\cref{fig:4}, previous pyramid-based methods for pixel-level prediction~\cite{ghiasi2016laplacian,chen2018reverse} started with \texttt{Stage-5}, and ended at \texttt{Stage-2}. 
        However, there are still two remaining stages to reconstruct for previous methods, which makes the reconstruction process incomplete in terms of the boundary quality. 
        Thus, we claim that starting image pyramid from \texttt{Stage-3} is sufficient, and should reconstruct until we encounter the lowest stage, \texttt{Stage-0} for HR results.
        To recover the scale of non-existing stages (\texttt{Stage-1, Stage-0}), we use bi-linear interpolation in appropriate locations (\cref{fig:4}).

        We locate a self-attention-based decoder, Scale Invariant Context Attention (SICA), on each stage to predict a Laplacian image of the saliency map (Laplacian saliency map). 
        From the predicted Laplacian saliency maps, we reconstruct saliency maps from higher-stages to the lower-stages (\cref{fig:4}a).
        \begin{figure}
    \centering
    \includegraphics[width=0.6\linewidth]{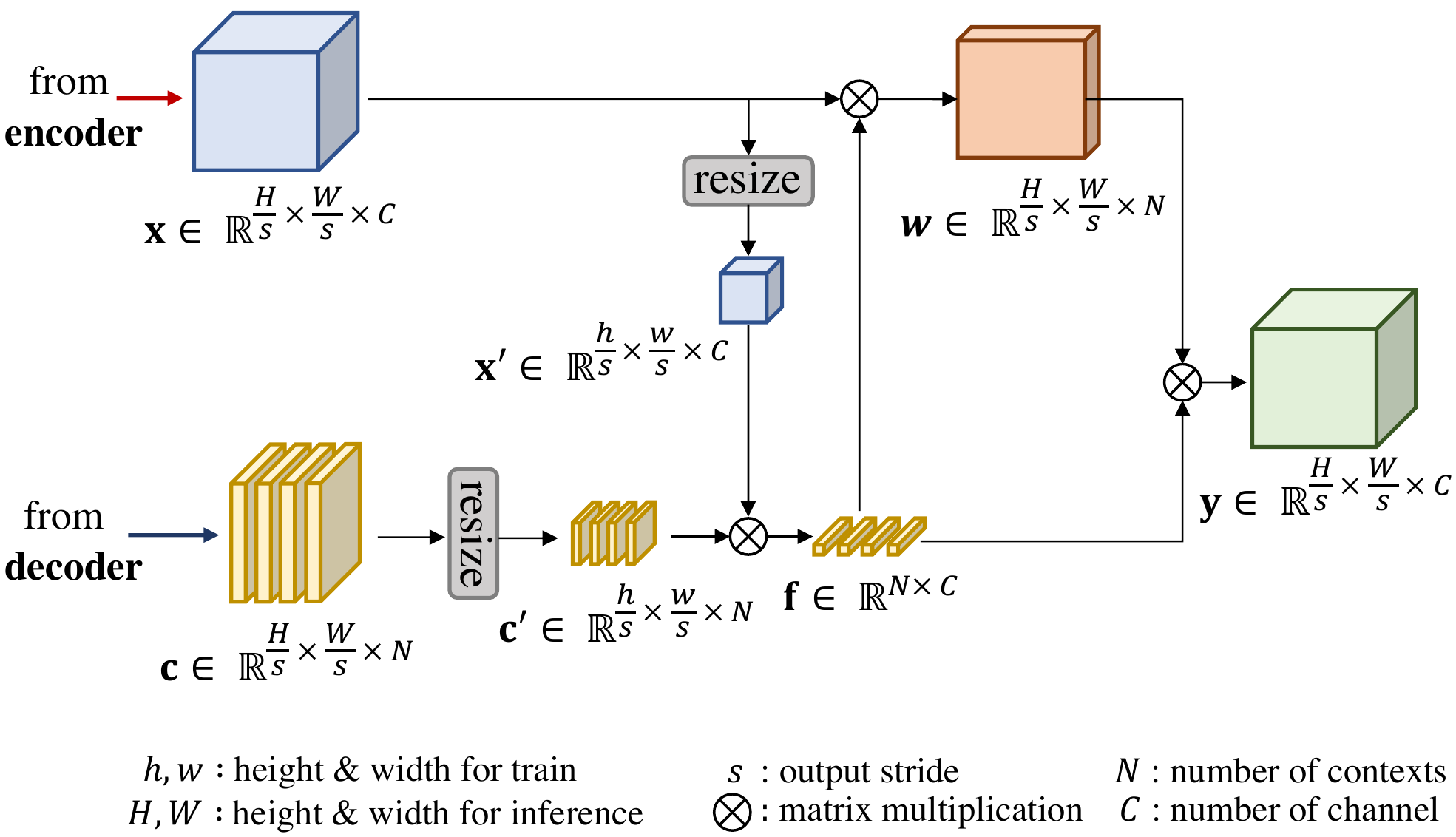}
    \caption{Illustration of Scale Invariant Context Attention (SICA)}
    \label{fig:5}
\end{figure}

        \noindent
        \textbf{Scale Invariant Context Attention.}
        \label{sec:met.2}
        Attention-based decoder for pixel-wise prediction shows great performance due to its non-local operation with respect to the spatial dimension~\cite{fu2019dual, yuan2020object}. 
        However, when the size of the input image gets larger than the training setting (\eg, $384 \times 384$), it usually fails to produce an appropriate result for the following reason.
        Because as the size of input image is large enough, there exist a train-inference discrepancies for a non-local operation which flattens the feature map according to the spatial dimension and does a matrix multiplication.
        For instance, the magnitude of the result from the non-local operation varies depending on the spatial dimension of the input image.
        Moreover, the complexity of non-local operation increases quadratically as the input size increases. 

        To this end, we propose SICA, a scale invariant context attention module for robust Laplacian saliency prediction. 
        As shown in~\cref{fig:5}, the overall operation of SICA follows OCRNet~\cite{yuan2020object}. 
        We found that computing object region representation causes train-inference discrepancy, so we resize input feature maps \textbf{x} and context maps \textbf{c} according to the shape from training time ($h, w$). 
        Because in the training step, images are already reshaped to the fixed shape, we do not have to resize them. 
        For context maps, unlike OCRNet, we can only access to the saliency map which is insufficient, so we generate several context maps following~\cite{kim2021uacanet}. 
        Further details of the context map selection and equations may be found in the supplementary material.
        With SICA, we can compute Laplacian saliency maps more precisely for HR images, and hence can apply pyramid blending for HR prediction (\cref{sec:met.5}).

        \noindent
        \textbf{Inverse Saliency Pyramid Reconstruction.}
        \label{sec:met.3}
        Laplacian pyramid~\cite{burt1983laplacian} is an image compression technique that stores the difference between the low-pass filtered image and the original image for each scale. 
        We can interpret the Laplacian image as a remainder from the low-pass filtered signal or, in other words, high-frequency details. 
        Inspired by this technique, we revisit the image pyramid structure by designing our network to construct a Laplacian pyramid to concentrate on the boundary details and reconstruct the saliency map from the smallest stage to its original size. 
        We start with the saliency map from the uppermost stage (\texttt{Stage-3}) for the initial saliency map and aggregate high-frequency details from the Laplacian saliency maps. 

        Formally, we denote the saliency map and Laplacian saliency map of the $j$th stage as $S^j$ and $U^j$, respectively. 
        To reconstruct the saliency map from the $j+1$th stage to the $j$th stage, we apply EXPAND operation~\cite{burt1983laplacian} as follows,
        \begin{equation}
            \label{eq:7}
            S^{j}_e(x,y) = 4 \sum_{m=-3}^{3} \sum_{n=-3}^{3} g(m,n) \cdot S^{j+1}(\frac{x-m}{2}, \frac{y-n}{2})
        \end{equation}
        where $(x, y)\in\mathcal{I}^j$ are pixel coordinates and $\mathcal{I}^j$ is a lattice domain of \texttt{Stage-j}. 
        Also, $g(m, n)$ is a Gaussian filter where the kernel size and standard deviation are empirically set to 7 and 1 respectively.
        To restore the saliency details, we add Laplacian saliency map from SICA as follows,
        \begin{equation}
            \label{eq:9}
            S^{j} = S^{j}_e + U^{j}.
        \end{equation}
        We repeat this process until we obtain the lowest stage, \texttt{Stage-0} as shown in~\cref{fig:4}a, and use it as a final prediction.
        
        \subsection{Supervision Strategy and Loss Functions}
        \label{sec:met.4}
        A typical way to supervise a network with multi-stage side outputs is to use bi-linear interpolation for each stage's prediction and compute the loss function with the ground-truth. 
        However, the predicted saliency map from higher-stage is small regarding its spatial dimension, and this may cause stage-scale inconsistency, especially for the boundary area of salient objects. 
        Instead, we focus on \textit{“Do what you can with what you have where you are''}.
        In fact, the saliency output from \texttt{Stage-3} cannot physically surpass the details from \texttt{Stage-2}, so we choose to provide each stage a suitable ground-truth. 
        To do so, we create an image pyramid of the ground-truth (\cref{fig:4}b).
        
        First, we obtain the ground-truth $G^j$ for \texttt{Stage-j} from $G^{j-1}$ with REDUCE operation~\cite{burt1983laplacian} as follows,
        \begin{equation}
            \label{eq:10}
            G^{j}(x,y) = \sum_{m=-3}^{3} \sum_{n=-3}^{3} g(m,n) \cdot G^{j-1}(2x+m, 2y+n).
        \end{equation}
        From the largest scale, we deconstruct the ground-truth until we get ground-truths for each stage of our network. 
        
        For loss function, we utilize binary cross entropy (BCE) loss with pixel position aware weighting strategy $\mathcal{L}^{wbce}$~\cite{wei2020f3net}.
        Moreover, to encourage the generated Laplacian saliency maps to follow the pyramid structure, we deconstruct $S^{j- 1}$ to the $j$th stage, $\tilde{S}^j$ by REDUCE operation. 
        Then, we reinforce the similarity between $S^j$ and reduced saliency map $\tilde{S}^j$ with pyramidal consistency loss $\mathcal{L}^{pc}$ as follows,
        \begin{equation}
            \label{eq:13}
            \mathcal{L}^{pc}(S^j, \tilde{S}^j) = \sum\limits_{(x,y) \in \mathcal{I}^j} || S^j(x, y) - \tilde{S}^j(x,y) ||_1.
        \end{equation}
        $\mathcal{L}^{pc}$ regularizes the lower-stage saliency maps to follow the structure of the image pyramid through the training process. 
        We define the total loss function $\mathcal{L}$ as follows,
        \begin{equation}
            \label{eq:14}
            \mathcal{L}(S,G) = \sum\limits^3_{j=0} \lambda_j \mathcal{L}^{wbce}(S^j, G^j) + \eta \sum\limits^2_{j=0} \lambda_j \mathcal{L}^{pc}(S^j, \tilde{S}^j)
        \end{equation}
        where $\eta$ is set to $10^{-4}$ and $\lambda_j = 4^j$ for balancing the magnitude of loss across stages.
        
        Finally, we include Stop-Gradient for the saliency map input of SICA and reconstruction process from higher-stages to force each stage saliency output to focus on each scale during training time and only affect each other in the inference time (\cref{fig:4}).
        This strategy encourages the stage-wise ground-truth scheme by explicitly preventing the gradient flow from lower-stages affecting the higher-stages. 
        Thus, supervisions with high-frequency details will not affect higher-stage decoder, which are intended only to have the abstract shape of the salient objects. 
        While this strategy might affect the performance in terms of the multiscale scheme, we use feature maps from the different stages for multiscale encoder and SICA to compensate for this issue.
        
        \begin{figure}
    \centering
    \includegraphics[width=0.6\linewidth]{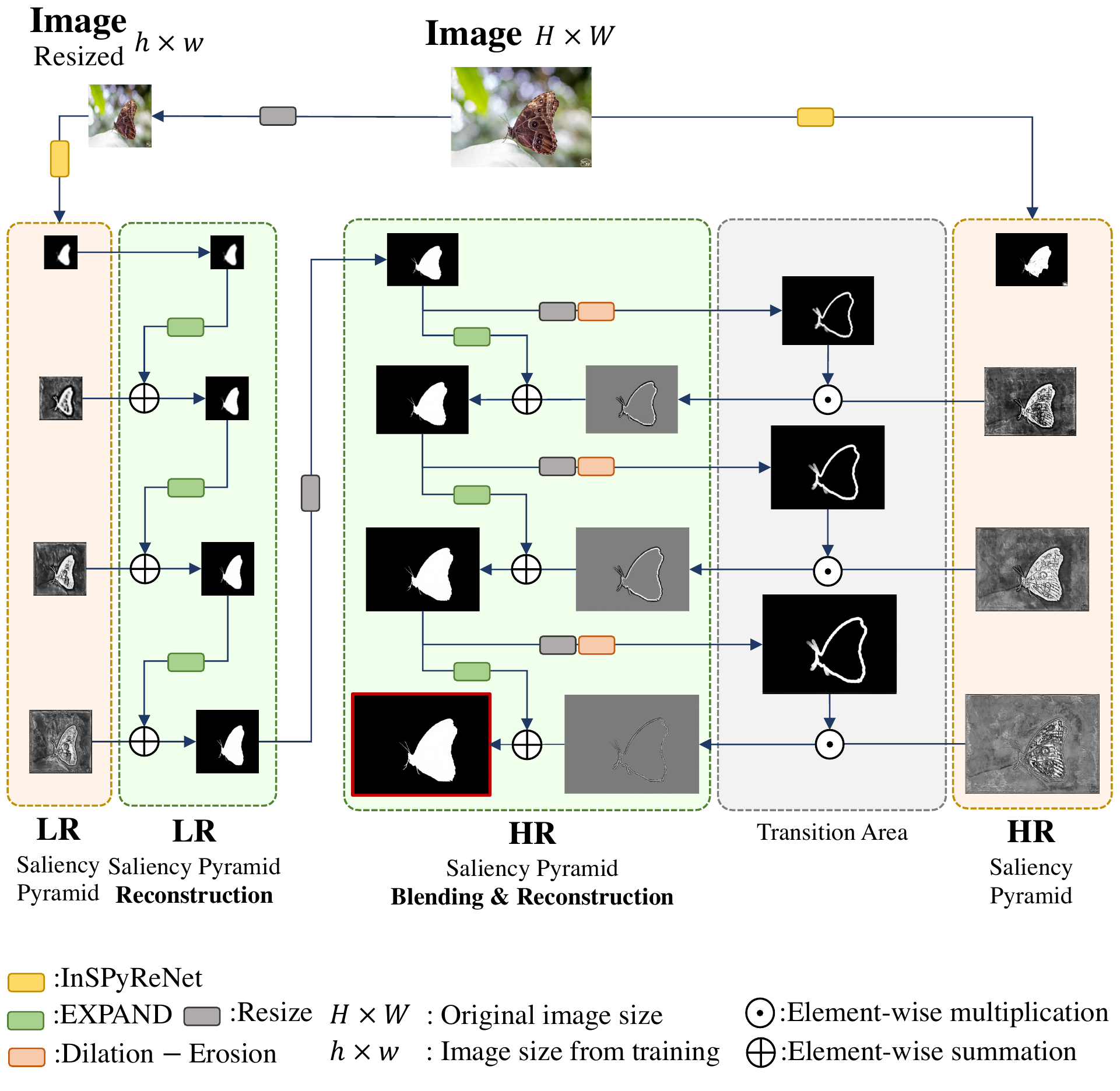}
    \caption{Illustration of pyramid blending of InSPyReNet for HR prediction.}
    \label{fig:6}
\end{figure}
        
        \subsection{Pyramid Blending}
        \label{sec:met.5}
        While SICA enables saliency prediction for various image sizes, when the image gets larger, there still exists ERF discrepancies (\cref{fig:3}).
        Thankfully, one very straightforward application for our saliency pyramid outputs is assembling multiple saliency pyramids from different inputs.
        We first generate saliency pyramids with InSPyReNet for original and resized images as shown in~\cref{fig:6}, namely LR and HR saliency pyramids.
        Then, instead of reconstructing the saliency map from the HR pyramid, we start from the lowest stage of the LR pyramid.
        Intuitively speaking, the LR pyramid is extended with the HR pyramid, so they construct a 7 stage saliency pyramid. 
        
        For the HR pyramid reconstruction, similar to~\cite{ghiasi2016laplacian}, we compute the dilation and erosion operation to the previous stage's saliency map and subtract them to obtain the transition area for and multiply with the Laplacian saliency map.
        Transition area is used to filter out the unwanted noises from the HR pyramid, since the boundary details we need to apply should exist only around the boundary area.
        Unlike~\cite{ghiasi2016laplacian}, it is unnecessary for the LR branch since we train InSPyReNet with methods in~\cref{sec:met.4}, results in the saliency pyramid are guaranteed to be consistent.
        

\begin{figure}
    \centering
    \includegraphics[width=0.7\textwidth]{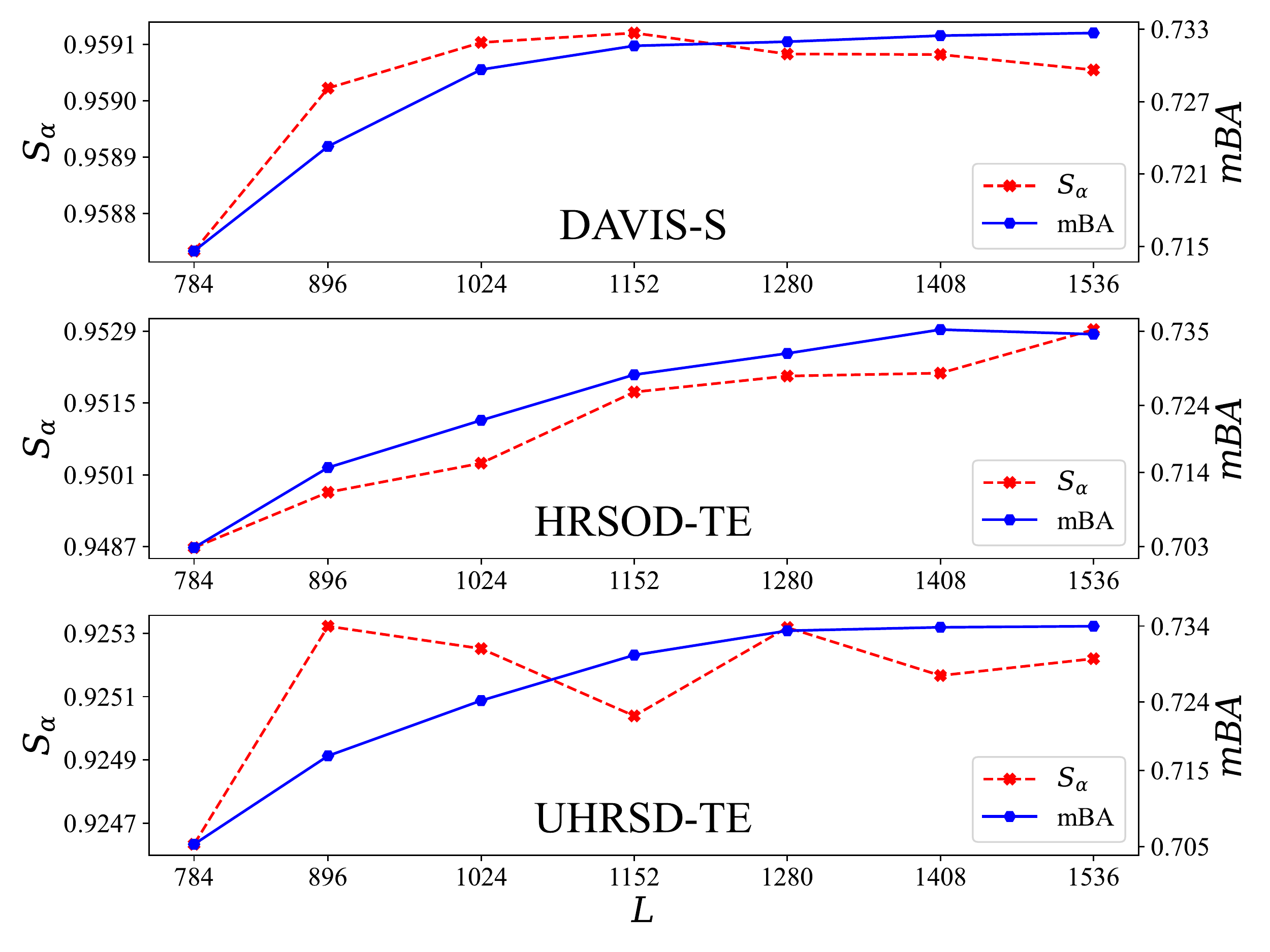}
    \caption{Performance measure ($S_\alpha$ and mBA) of InSPyReNet with pyramid blending by changing $L$ on three HR benchmarks.}
    \label{fig:7}
\end{figure}
\section{Experiments and Results}
\label{sec:exp}
\subsection{Experimental Settings}
    \noindent
    \textbf{Implementation Details.} We train our method with widely used DUTS-TR, a subset of DUTS~\cite{wang2017learning} for training. 
    We use Res2Net~\cite{gao2019res2net} or Swin Transformer~\cite{liu2021swin} backbone which is pre-trained with ImageNet-1K or ImageNet-22K~\cite{russakovsky2015imagenet} respectively. 
    Images are resized to $384 \times 384$ for training, and we use a random scale in a range of [0.75, 1.25] and crop to the original size, random rotation from -10 to 10 degrees, and random image enhancement (contrast, sharpness, brightness) for the data augmentation. 
    We set the batch size to 6 and maximum epochs to 60. 
    We use Adam optimizer~\cite{kingma2014adam} with initial learning rate 1e-5, and follow the default PyTorch settings. 
    Finally, we use poly learning rate decay for scheduling~\cite{zhao2017pyramid} with a factor of $(1-(\frac{iter}{iter_{\max}})^{0.9})$ and linear warm-up for the first 12000 iterations.

    \noindent
    \textbf{Evaluation Datasets and Metrics.} We evaluate our method on five LR benchmarks, DUTS-TE, a subset of DUTS for evaluation, DUT-OMRON~\cite{yang2013saliency}, ECSSD~\cite{shi2015hierarchical}, HKU-IS~\cite{li2015visual}, and PASCAL-S~\cite{li2014secrets}.
    Furthermore, we evaluate our method on three HR benchmarks, DAVIS-S~\cite{perazzi2016benchmark}, HRSOD-TE~\cite{zeng2019towards}, and UHRSD-TE~\cite{xie2022pyramid}.
    From~\cite{wang2021salient}, we report S-measure ($S_\alpha$)~\cite{fan2017structure}, maximum F-measure ($F_{\max}$)~\cite{achanta2009frequency}, and Mean Absolute Error (MAE)~\cite{perazzi2012saliency}. 
    Since F-measure requires a binary map, it is computed with thresholds in a range of [0, 255] and the maximum value is used for the evaluation. 
    With the above metrics, we also report mean boundary accuracy (mBA)~\cite{cheng2020cascadepsp} for boundary quality measure.
    
    \subsection{Ablation Studies}
    \label{sec:exp.4}
    \begin{table}
    \setlength{\tabcolsep}{2.5pt}
    \centering
    \caption{Ablation study of InSPyReNet (SwinB) with and without SICA and pyramid blending on HR benchmarks.}
    \resizebox{\linewidth}{!}{
        \begin{tabular}{c|c|cccc|cccc|cccc}
            \hline \hline
            \multirow{2}{*}{Resolution} & \multirow{2}{*}{Setting} & \multicolumn{4}{c|}{DAVIS-S} & \multicolumn{4}{c|}{HRSOD-TE} & \multicolumn{4}{c}{UHRSD-TE} \\
            & & $S_\alpha\uparrow$ & $F_{\max}\uparrow$ & MAE$\downarrow$ & mBA$\uparrow$ & 
              $S_\alpha\uparrow$ & $F_{\max}\uparrow$ & MAE$\downarrow$ & mBA$\uparrow$ & 
              $S_\alpha\uparrow$ & $F_{\max}\uparrow$ & MAE$\downarrow$ & mBA$\uparrow$ \\ \hline\hline
              \multicolumn{14}{c}{w/o pyramid blending} \\ \hline\hline
            $384 \times 384$   &      -    & 0.953 & 0.949 & 0.013 & 0.705 & 0.945 & 0.941 & 0.019 & 0.700 & 0.927 & 0.932 & 0.032 & 0.724 \\
            $L=1280$ & w/o SICA  & 0.396 & 0.602 & 0.497 & 0.504 & 0.373 & 0.416 & 0.530 & 0.512 & 0.242 & 0.395 & 0.645 & 0.506 \\
            $L=1280$ & w/ SICA   & 0.873 & 0.821 & 0.037 & \textbf{0.774} & 0.886 & 0.873 & 0.043 & \textbf{0.750} & 0.809 & 0.819 & 0.092 & \textbf{0.751} \\
            \hline\hline
            \multicolumn{14}{c}{w/ pyramid blending} \\ \hline\hline
            $L=1280$ & w/o SICA & 0.860 & 0.883 & 0.023 & 0.537 & 0.863 & 0.869 & 0.029 & 0.531 & 0.834 & 0.863 & 0.052 & 0.521 \\
            \rowgray
            $L=1280$ & w/ SICA  & \textbf{0.962} & \textbf{0.959} & \textbf{0.009} & 0.743 & \textbf{0.952} & \textbf{0.949} & \textbf{0.016} & 0.738 & \textbf{0.932} & \textbf{0.938} & \textbf{0.029} & 0.741 \\
            \hline \hline
            
        \end{tabular}}
    \label{tab:1}
\end{table}
    
    \noindent
    \textbf{Resizing Factor \textit{L}.} We use the resizing method for HR images from~\cite{cheng2020cascadepsp} for pyramid blending since current GPUs cannot deal with large sizes such as 4K images as is. 
    So, we choose a maximum length of the shorter side of the image as $L$. For instance, if an input size is $1920 \times 1080$ and $L=810$, then we resize the input into $1440 \times 810$.
    Moreover, we do not deploy pyramid blending process for inputs where the shorter side length is less than 512 because the difference between LR and HR pyramid is not enough for blending.
    We compare $S_\alpha$ and mBA on three HR datasets by varying $L$ from 784 to 1536 (\cref{fig:7}). We choose $L=1280$ since mBA almost converges after that.
    
    \noindent
    \textbf{SICA and pyramid blending.} To demonstrate the necessity of SICA, we evaluate InSPyReNet with and without SICA.
    Since SICA only takes place when the input image is large enough to make train-inference discrepancy, we demonstrate results only on HR benchmarks. 
    Please note that all evaluation is done with resizing method mentioned above, except for the LR resolution (\cref{tab:1}).
    
    In~\cref{tab:1}, InSPyReNet without SICA shows the worst performance, especially for the mBA. 
    Since mBA only considers boundary quality, InSPyReNet with SICA and without pyramid blending shows the best performance in terms of mBA measure, yet shows poor results on other SOD metrics\footnote{This phenomenon shows that mBA itself cannot measure the performance of saliency detection, rather it only measures the quality of boundary itself.}.
    This is because even with SICA, InSPyReNet cannot overcome the discrepancy in effective receptive fields~\cite{luo2016understanding} between HR and LR images.
    For the setting without SICA, InSPyReNet with pyramid blending shows inferior results compared to the InSPyReNet without pyramid blending, meaning that the pyramid blending technique is meaningless without SICA since it worsen the results.
    Thus, SICA is crucial to be included in InSPyReNet, especially for the HR pyramid in the pyramid blending.
    Compared to the LR setting (\ie, resizing into $384 \times 384$), using both SICA and pyramid blending shows better performance for all four metrics.
    
    \subsection{Comparison with State-of-the-Art methods} 
    \begin{table}
    \setlength{\tabcolsep}{2.5pt}
    \centering
    \caption{Quantitative results on five LR benchmarks. 
    The first and the second best results for each metric are colored \textcolor{red}{red} and \textcolor{blue}{blue}.
        $\uparrow$ indicates larger the better, and $\downarrow$ indicates smaller the better.
        $\dagger$ indicates our re-implementation.}
    \resizebox{\textwidth}{!}{
        \begin{tabular}{c|c|ccc|ccc|ccc|ccc|ccc}
            \hline \hline
            \multirow{2}{*}{Algorithms}
            & \multirow{2}{*}{Backbones}
            & \multicolumn{3}{c|}{DUTS-TE}             
            & \multicolumn{3}{c|}{DUT-OMRON}                                   
            & \multicolumn{3}{c|}{ECSSD}                                       
            & \multicolumn{3}{c|}{HKU-IS}                                 
            & \multicolumn{3}{c}{PASCAL-S} \\
            & 
            & $S_\alpha\uparrow$ & $F_{\max}\uparrow$ & MAE$\downarrow$
            & $S_\alpha\uparrow$ & $F_{\max}\uparrow$ & MAE$\downarrow$
            & $S_\alpha\uparrow$ & $F_{\max}\uparrow$ & MAE$\downarrow$
            & $S_\alpha\uparrow$ & $F_{\max}\uparrow$ & MAE$\downarrow$
            & $S_\alpha\uparrow$ & $F_{\max}\uparrow$ & MAE$\downarrow$\\ \hline \hline
            \multicolumn{17}{c}{CNN backbone Models (ResNet, ResNext, Res2Net)} \\
            \hline \hline
            PoolNet~\cite{liu2019simple}                 & ResNet50 & 0.887 & 0.865 & 0.037 & 0.831 & 0.763 & 0.054 & 0.926 & 0.937 & 0.035 & 0.909 & 0.912 & 0.034 & 0.865 & 0.858 & 0.065 \\ 
            BASNet~\cite{qin2019basnet}                  & ResNet34 & 0.866 & 0.838 & 0.048 & 0.836 & 0.779 & 0.056 & 0.916 & 0.931 & 0.037 & 0.909 & 0.919 & 0.032 & 0.838 & 0.835 & 0.076 \\ 
            EGNet~\cite{zhao2019egnet}                   & ResNet50 & 0.874 & 0.848 & 0.045 & 0.836 & 0.773 & 0.057 & 0.918 & 0.928 & 0.041 & 0.915 & 0.920 & 0.032 & 0.848 & 0.836 & 0.075 \\ 
            CPD~\cite{wu2019cascaded}                    & ResNet50 & 0.869 & 0.840 & 0.043 & 0.825 & 0.754 & 0.056 & 0.918 & 0.926 & 0.037 & 0.905 & 0.911 & 0.034 & 0.848 & 0.833 & 0.071 \\ 
            GateNet~\cite{zhao2020suppress}              & ResNeXt101 & 0.897 & 0.880 & 0.035 & \textcolor{blue}{0.849} & \textcolor{blue}{0.794} & 0.051 & 0.929 & 0.940 & 0.035 & \textcolor{blue}{0.925} & 0.932 & 0.029 & 0.865 & 0.855 & 0.064 \\ 
            $^\dagger$Chen \etal~\cite{chen2018reverse}         & Res2Net50 & 0.890 & 0.869 & 0.040 & 0.834 & 0.769 & 0.061 & \textcolor{blue}{0.931} & \textcolor{blue}{0.943} & 0.035 & 0.921 & 0.927 & 0.034 & \textcolor{blue}{0.871} & 0.862 & 0.060 \\ 
            $^\dagger$F$^3$Net~\cite{wei2020f3net}       & Res2Net50 & 0.892 & 0.876 & \textcolor{blue}{0.033} & 0.839 & 0.771 & \textcolor{blue}{0.048} & 0.915 & 0.925 & 0.040 & 0.915 & 0.925 & 0.030 & 0.856 & 0.842 & 0.065 \\ 
            $^\dagger$LDF~\cite{wei2020label}            & Res2Net50 & 0.897 & 0.885 & \textcolor{red}{0.032} & 0.848 & 0.788 & \textcolor{red}{0.045} & 0.928 & \textcolor{blue}{0.943} & 0.033 & 0.924 & \textcolor{blue}{0.935} & \textcolor{red}{0.027} & 0.868 & \textcolor{blue}{0.863} & \textcolor{blue}{0.059} \\ 
            $^\dagger$MINet~\cite{pang2020multi}         & Res2Net50 & 0.896 & 0.883 & 0.034 & 0.843 & 0.787 & 0.055 & \textcolor{blue}{0.931} & 0.942 & \textcolor{red}{0.031} & 0.923 & 0.931 & \textcolor{blue}{0.028} & 0.865 & 0.858 & 0.060 \\ 
            $^\dagger$PA-KRN~\cite{xu2021locate}         & Res2Net50 & \textcolor{blue}{0.898} & \textcolor{blue}{0.888} & 0.034 & \textcolor{red}{0.853} & \textcolor{red}{0.808} & 0.050 & 0.930 & \textcolor{blue}{0.943} & \textcolor{blue}{0.032} & 0.922 & \textcolor{blue}{0.935} & \textcolor{red}{0.027} & 0.863 & 0.859 & 0.063 \\ 
            \hline
            \rowgray
            \textit{Ours}                                & Res2Net50 & \textcolor{red}{0.904} & \textcolor{red}{0.892} & 0.035 & 0.845 & 0.791 & 0.059 & \textcolor{red}{0.936} & \textcolor{red}{0.949} & \textcolor{red}{0.031} & \textcolor{red}{0.929} & \textcolor{red}{0.938} & \textcolor{blue}{0.028} & \textcolor{red}{0.876} & \textcolor{red}{0.869} & \textcolor{red}{0.056} \\ 
            \hline \hline
            \multicolumn{17}{c}{Transformer backbone Models (Swin, T2T-ViT)} \\
            \hline\hline
            VST~\cite{liu2021visual}                     & T2T-ViT-14 & 0.896 & 0.878 & 0.037 & 0.850 & 0.800 & 0.058 & 0.932 & 0.944 & 0.033 & 0.928 & 0.937 & 0.029 & 0.872 & 0.864 & 0.061 \\ 
            Mao \etal~\cite{mao2021transformer}              & SwinB & \textcolor{blue}{0.917} & \textcolor{blue}{0.911} & \textcolor{blue}{0.025} & 0.862 & 0.818 & 0.048 & \textcolor{blue}{0.943} & \textcolor{blue}{0.956} & \textcolor{red}{0.022} & \textcolor{blue}{0.934} & \textcolor{blue}{0.945} & \textcolor{blue}{0.022} & \textcolor{blue}{0.883} & \textcolor{blue}{0.883} & \textcolor{blue}{0.050} \\ 
            $^\dagger$Chen \etal~\cite{chen2018reverse}         & SwinB & 0.901 & 0.883 & 0.034 & 0.860 & 0.810 & 0.052 & 0.937 & 0.948 & 0.030 & 0.928 & 0.935 & 0.029 & 0.876 & 0.868 & 0.058 \\ 
            $^\dagger$F$^3$Net~\cite{wei2020f3net}       & SwinB & 0.902 & 0.895 & 0.033 & 0.860 & 0.826 & 0.053 & 0.937 & 0.951 & 0.027 & 0.932 & 0.944 & 0.023 & 0.868 & 0.864 & 0.059 \\ 
            $^\dagger$LDF~\cite{wei2020label}            & SwinB & 0.896 & 0.881 & 0.036 & 0.854 & 0.809 & 0.052 & 0.931 & 0.942 & 0.032 & 0.933 & 0.941 & 0.024 & 0.861 & 0.851 & 0.065 \\ 
            $^\dagger$MINet~\cite{pang2020multi}         & SwinB & 0.906 & 0.893 & 0.029 & 0.852 & 0.798 & 0.047 & 0.935 & 0.949 & 0.028 & 0.930 & 0.938 & 0.025 & 0.875 & 0.870 & 0.054 \\ 
            $^\dagger$PA-KRN~\cite{xu2021locate}         & SwinB & 0.913 & 0.906 & 0.028 & \textcolor{blue}{0.874} & \textcolor{red}{0.838} & \textcolor{red}{0.042} & 0.941 & \textcolor{blue}{0.956} & 0.025 & 0.933 & 0.944 & 0.023 & 0.873 & 0.872 & 0.056 \\ 
            \hline
            \rowgray
            \textit{Ours}                                & SwinB & \textcolor{red}{0.931} & \textcolor{red}{0.927} & \textcolor{red}{0.024} & \textcolor{red}{0.875} & \textcolor{blue}{0.832} & \textcolor{blue}{0.045} & \textcolor{red}{0.949} & \textcolor{red}{0.960} & \textcolor{blue}{0.023} & \textcolor{red}{0.944} & \textcolor{red}{0.955} & \textcolor{red}{0.021} & \textcolor{red}{0.893} & \textcolor{red}{0.893} & \textcolor{red}{0.048} \\ 
            \hline\hline

        \end{tabular}}
    \label{tab:2}
\end{table}
    
    \label{sec:exp.3}
    \noindent
    \textbf{Quantitative Comparison.} First, we compare InSPyReNet with 12 SotA LR SOD methods. In this experiment, we resize images same as for training.
    We either download pre-computed saliency maps or run an official implementation with pre-trained model parameters provided by the authors to evaluate with the same evaluation code for a fair comparison. 
    Moreover, we re-implement Chen \etal~\cite{chen2018reverse}, F$^3$Net~\cite{wei2020f3net}, LDF~\cite{wei2020label}, MINet~\cite{pang2020multi}, and PA-KRN~\cite{xu2021locate} with same backbones we use to demonstrate how much the training settings affects the performance for other methods compared to InSPyReNet.
    We choose the above methods since they provided source code with great reproducibility and consistent results.
    As shown in~\cref{tab:2}, our SwinB backbone model consistently shows outstanding performance across three metrics. 
    Moreover, our Res2Net50 backbone model shows competitive results regarding its number of parameters.
    
    Moreover, to verify the effectiveness of pyramid blending, we compare our method with SotA methods on HR and LR benchmarks (\cref{tab:3}).
    Among HR methods, our method shows great performance among other methods, even though we use only DUTS-TR for training. 
    Note that previous SotA HR methods show inferior results on LR datasets and vice versa, meaning that generalizing for both scales is difficult, while our method is robust for both scales.
    For instance, while PGNet trained with HR datasets (H, U) shows great performance on HR benchmarks, 
    but shows more inferior results than other methods and even LR methods on LR benchmarks, while our method shows consistent results on both benchmarks.
    This is because LR datasets do not provide high-quality boundary details, while HR datasets lack of global object saliency.
    
    \begin{table}
    \setlength{\tabcolsep}{2.5pt}
    \centering
    \caption{Quantitative results on three HR and two LR benchmarks. 
    Beckbones; V: VGG16, R18: ResNet18, R50: ResNet50, S: SwinB. Datasets; D: DUTS-TR, H: HRSOD-TR, U: UHRSD-TR. 
    The first and the second best results for each metric are colored \textcolor{red}{red} and \textcolor{blue}{blue}. 
    $\uparrow$ indicates larger the better, and $\downarrow$ indicates smaller the better.
    $\dagger$ indicates our re-implementation.}
    \resizebox{\linewidth}{!}{
        \begin{tabular}{c|c|c|cccc|cccc|cccc|ccc|ccc}
            \hline \hline
            \multirow{3}{*}{Algorithms} & \multirow{3}{*}{Backbone} & \multirow{3}{*}{\shortstack{Train\\Datasets}} & \multicolumn{12}{c|}{HR benchmarks} & \multicolumn{6}{c}{LR benchmarks} \\ \cline{4-21}
            & & & \multicolumn{4}{c|}{DAVIS-S} & \multicolumn{4}{c|}{HRSOD-TE} & \multicolumn{4}{c|}{UHRSD-TE} & \multicolumn{3}{c|}{DUTS-TE} & \multicolumn{3}{c}{DUT-OMRON}  \\
            & & & $S_\alpha\uparrow$ & $F_{\max}\uparrow$ & MAE$\downarrow$ & mBA$\uparrow$ & $S_\alpha\uparrow$ & $F_{\max}\uparrow$ & MAE$\downarrow$ & mBA$\uparrow$ & $S_\alpha\uparrow$ & $F_{\max}\uparrow$ & MAE$\downarrow$ & mBA$\uparrow$ & $S_\alpha\uparrow$ & $F_{\max}\uparrow$ & MAE$\downarrow$ & $S_\alpha\uparrow$ & $F_{\max}\uparrow$ & MAE$\downarrow$ \\ \hline\hline
            $^\dagger$Chen \etal~\cite{chen2018reverse}           & S & D & 0.934 & 0.925 & 0.018 & 0.697 & 0.915 & 0.907 & 0.032 & 0.684 & 0.915 & 0.919 & 0.034 & 0.712 & 0.901 & 0.883 & 0.034 & 0.860 & 0.810 & 0.052 \\
            $^\dagger$F$^3$Net~\cite{wei2020f3net}         & S & D & 0.931 & 0.922 & 0.017 & 0.681 & 0.912 & 0.902 & 0.034 & 0.674 & 0.920 & 0.922 & 0.033 & 0.708 & 0.902 & 0.895 & 0.033 & 0.860 & 0.826 & 0.053 \\
            $^\dagger$LDF~\cite{wei2020label}              & S & D & 0.928 & 0.918 & 0.019 & 0.682 & 0.905 & 0.888 & 0.036 & 0.672 & 0.911 & 0.913 & 0.038 & 0.702 & 0.896 & 0.881 & 0.036 & 0.854 & 0.809 & 0.052 \\
            $^\dagger$MINet~\cite{pang2020multi}           & S & D & 0.933 & 0.930 & 0.017 & 0.673 & 0.927 & 0.917 & 0.025 & 0.670 & 0.915 & 0.917 & 0.035 & 0.694 & 0.906 & 0.893 & 0.029 & 0.852 & 0.798 & 0.047 \\
            $^\dagger$PA-KRN~\cite{xu2021locate}           & S & D & 0.944 & 0.935 & 0.014 & 0.668 & 0.927 & 0.918 & 0.026 & 0.653 & 0.919 & 0.926 & 0.034 & 0.673 & \textcolor{blue}{0.913} & \textcolor{blue}{0.906} & 0.028 & \textcolor{blue}{0.874} & \textcolor{red}{0.838} & \textcolor{red}{0.042} \\
            PGNet~\cite{xie2022pyramid}          & S+R18 & D   & 0.935 & 0.931 & 0.015 & 0.707 & 0.930 & 0.922 & 0.021 & 0.693 & 0.912 & 0.914 & 0.037 & 0.715 & 0.911 & 0.903 & \textcolor{blue}{0.027} & 0.855 & 0.803 & \textcolor{blue}{0.045} \\ 
            Zeng \etal~\cite{zeng2019towards}         & V & D,H & 0.876 & 0.889 & 0.026 & 0.618 & 0.897 & 0.892 & 0.030 & 0.623 & - & - & - & - & 0.824 & 0.835 & 0.051 & 0.762 & 0.743 & 0.065 \\
            Tang \etal~\cite{tang2021disentangled}   & R50 & D,H & 0.920 & 0.935 & 0.012 & 0.716 & 0.920 & 0.915 & 0.022 & 0.693 & - & - & - & - & 0.895 & 0.888 & 0.031 & 0.843 & 0.796 & 0.048 \\
            PGNet~\cite{xie2022pyramid}          & S+R18 & D,H & 0.947 & 0.948 & 0.012 & 0.716 & 0.935 & 0.929 & \textcolor{blue}{0.020} & 0.714 & 0.912 & 0.915 & 0.036 & 0.735 & 0.912 & 0.905 & 0.028 & 0.858 & 0.803 & 0.046 \\ 
            PGNet~\cite{xie2022pyramid}          & S+R18 & H,U & \textcolor{blue}{0.954} & \textcolor{blue}{0.956} & \textcolor{blue}{0.010} & \textcolor{blue}{0.730} & \textcolor{blue}{0.938} & \textcolor{blue}{0.939} & \textcolor{blue}{0.020} & \textcolor{blue}{0.727} & \textcolor{red}{0.935} & \textcolor{blue}{0.930} & \textcolor{red}{0.026} & \textcolor{red}{0.765} & 0.861 & 0.828 & 0.038 & 0.790 & 0.727 & 0.059 \\ 
            \hline
            \rowgray
            \textit{Ours} & S & D   & \textcolor{red}{0.962} & \textcolor{red}{0.959} & \textcolor{red}{0.009} & \textcolor{red}{0.743} & \textcolor{red}{0.952} & \textcolor{red}{0.949} & \textcolor{red}{0.016} & \textcolor{red}{0.738} & \textcolor{blue}{0.932} & \textcolor{red}{0.938} & \textcolor{blue}{0.029} & \textcolor{blue}{0.741} & \textcolor{red}{0.931} & \textcolor{red}{0.927} & \textcolor{red}{0.024} & \textcolor{red}{0.875} & \textcolor{blue}{0.832} & \textcolor{blue}{0.045} \\
            \hline \hline
        \end{tabular}}
    \label{tab:3}
\end{table}
    
    \begin{figure}
    \centering
    \includegraphics[width=\linewidth]{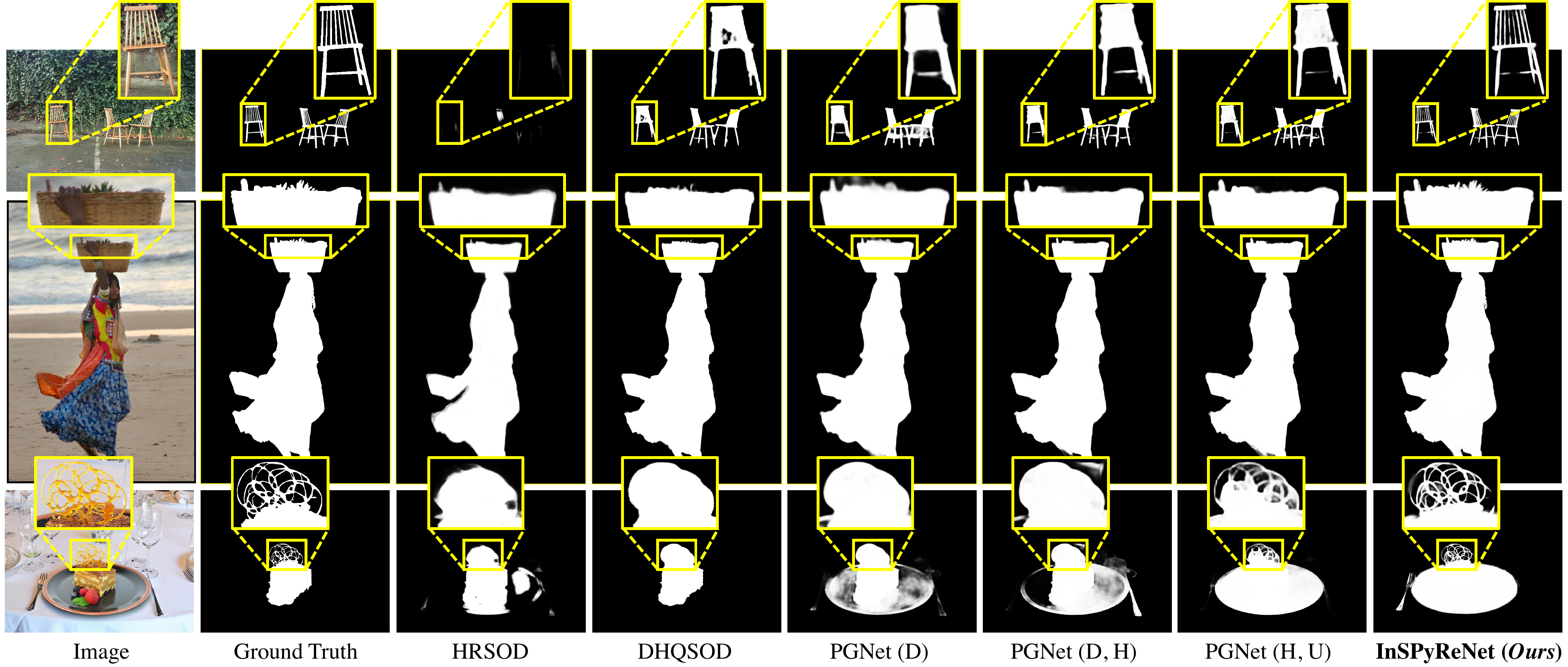} 
    \caption{Qualitative results of InSPyReNet (SwinB) compared to SotA HR methods on HRSOD-TE. \textit{Best viewed by zooming in.}}
    \label{fig:8}
\end{figure}
    \begin{figure}
    \centering
    \includegraphics[width=0.8\linewidth]{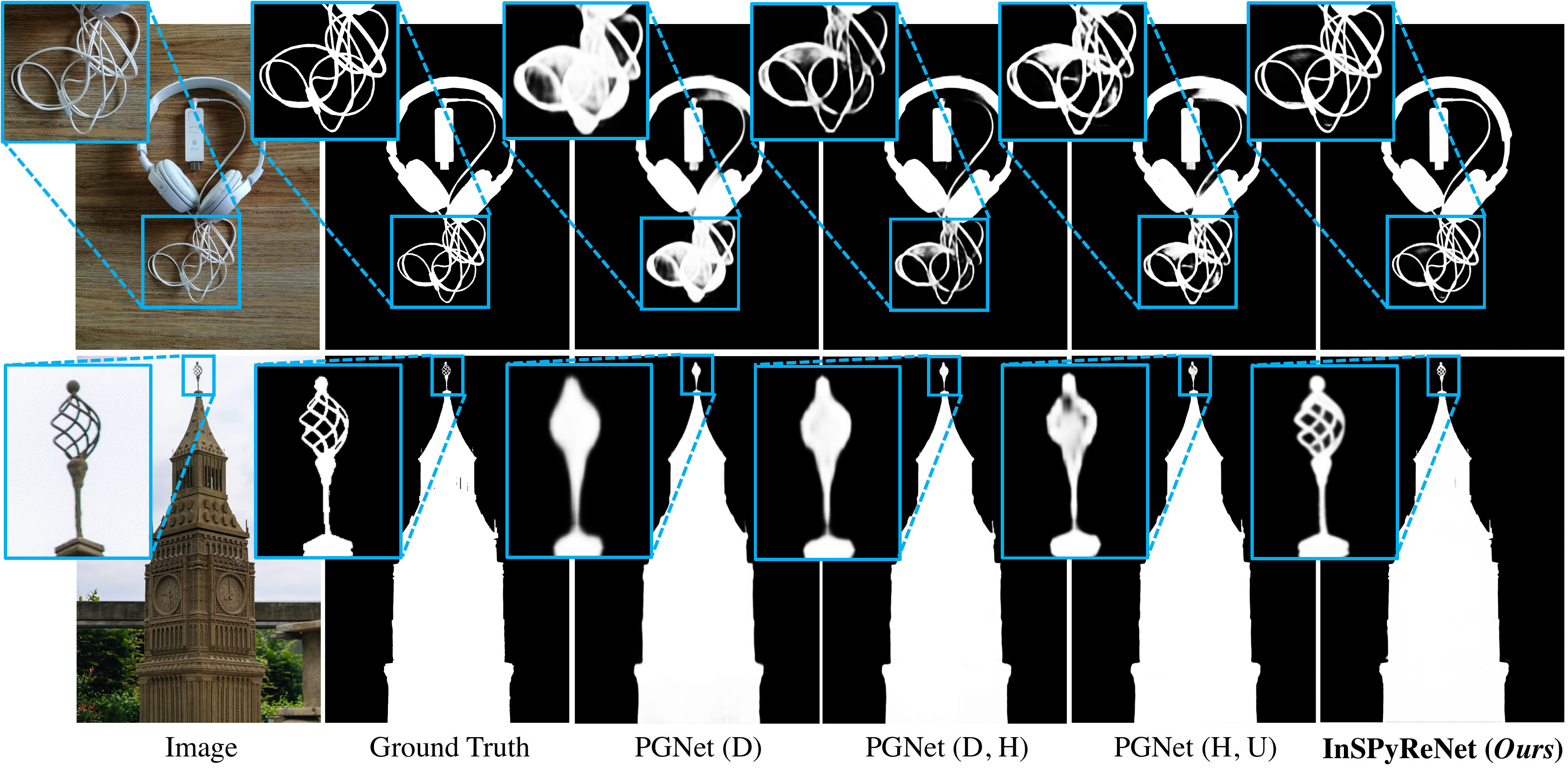} 
    \caption{Qualitative results of InSPyReNet (SwinB) compared to PGNet on UHRSD-TE. \textit{Best viewed by zooming in.}}
    \label{fig:9}
\end{figure}
    \begin{figure}
    \centering
    \includegraphics[width=\linewidth]{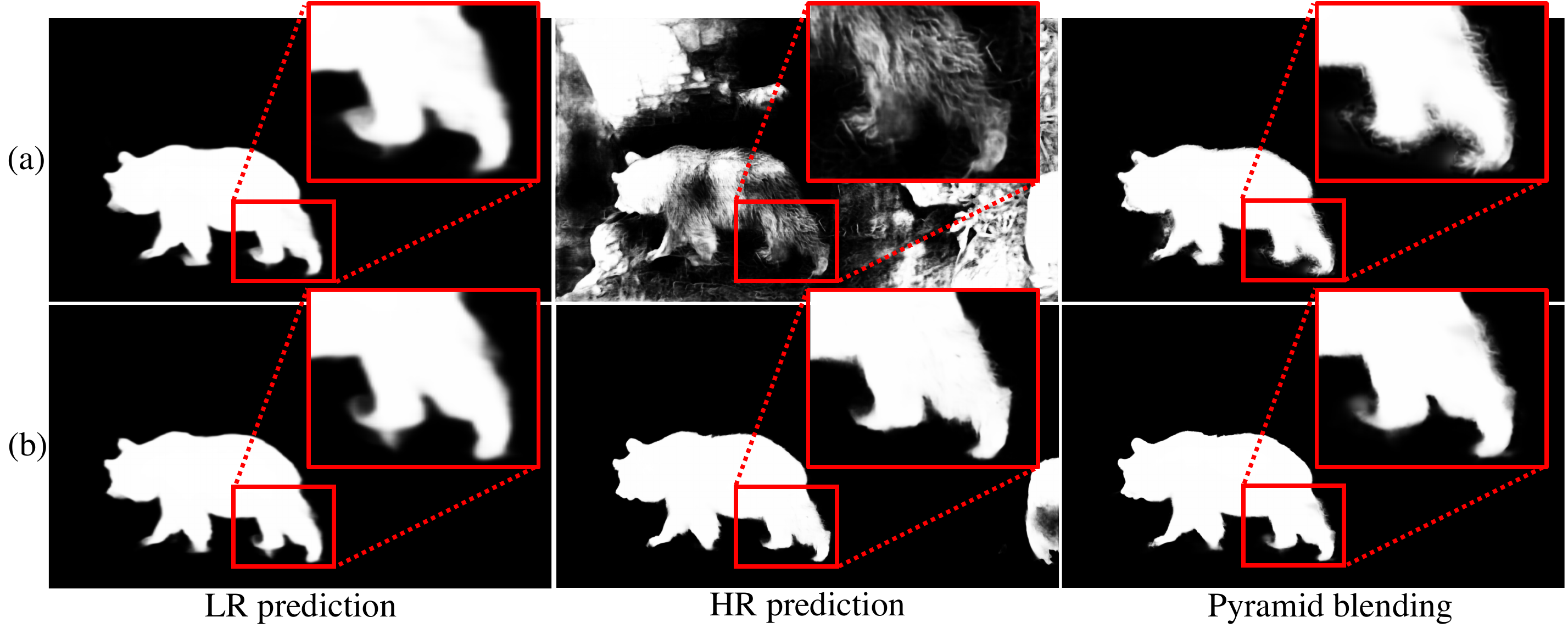} 
    \caption{Visual comparison of LR, HR prediction, and pyramid blended results of InSPyReNet with (a) Res2Net50 and (b) SwinB backbones.}
    \label{fig:10}
\end{figure}

    \noindent
    \textbf{Qualitative Comparison.}
    We provide a visual comparison of our method in~\cref{fig:8} and~\cref{fig:9} on HR benchmarks. 
    Overall, previous SotA methods are sufficient for detecting salient objects, but shows degraded results for complex scenes.
    Results show that InSPyReNet can produce accurate saliency prediction for the complex, fine details thanks to the pyramid blending.
    Moreover, even though we train our method only with LR dataset, DUTS-TR, InSPyReNet consistently shows accurate results compared to other methods.

\section{Discussion and Conclusion}
\label{sec:disc}

    \noindent
    \textbf{Weakness: Backbone Network.} We do not use Res2Net50 for HR benchmark due to the following reason. 
    As shown in~\cref{fig:10}, HR prediction from Res2Net50 backbone produces saliency map with numerous unnecessary artifacts.
    This is because CNN backbones are vulnerable to its ERF size, which is highly dependent on its training dataset.
    Unlike traditional CNN backbones, there are many works to minimize the above issue such as Fast Fourier Convolution~\cite{chi2020ffc}, or ConvNeXt~\cite{liu2022convnet}.
    We found that those methods are helpful for reducing such artifacts for HR prediction, but not enough for detail reconstruction.
    However, Vision Transformers like SwinB have larger ERFs and consist of non-local operation for regarding global dependencies, which are suitable for our method.
    Thus, even the HR prediction shows some False Positives (second column, second row in~\cref{fig:10}), we can easily remove them while enhance boundary details via pyramid blending. 

    \noindent
    \textbf{Future Work and Conclusion.} Starting from previous works with Laplacian pyramid prediction~\cite{ghiasi2016laplacian,chen2018reverse}, 
    we have shown that InSPyReNet shows noticeable improvements on HR prediction without any HR training datasets or complex architecture.
    In a series of experiments, our method shows great performance on HR benchmarks while robust again LR benchmarks as well.
    Although we only utilize a concept of pyramid-based image blending for merging two pyramids with different scales,
    we hope our work can extend to the multi-modal input such as RGB-D SOD or video SOD with temporal information.

\noindent
\textbf{Acknowledgement.}
This work was supported by Institute of Information \& communications Technology Planning \& Evaluation(IITP) grant funded by the Korea government(MSIT) 
(No.2017-0-00897, Development of Object Detection and Recognition for Intelligent Vehicles) and 
(No.B0101-15-0266, Development of High Performance Visual BigData Discovery Platform for Large-Scale Realtime Data Analysis)

\bibliographystyle{splncs}
\bibliography{egbib}

\appendix
\phantomsection
\addcontentsline{toc}{chapter}{Appendices}
\chapter*{Supplementary Material for Revisiting Image Pyramid Structure for High Resolution Salient Object Detection}
\setcounter{section}{0}
\renewcommand{\thesection}{\Alph{section}}

\section{Method Details}

\subsection{Backbone Networks}
In this section, we describe how the feature maps from the backbone network is retrieved. 
Since we adopt two different types of backbone networks, we explain the minor details of each model.

For Res2Net~\cite{gao2019res2net} backbone network, we use 26w$\times$8s setting.
Because the main difference between ResNet~\cite{he2016deep} and Res2Net is a building block, the overall architecture is identical.
So, to explain where we extract the feature maps from the Res2Net backbone, we refer to the ResNet paper. 
The feature map for the \texttt{Stage-1} is extracted from conv1, and for \texttt{Stage-j}, where $j > 1$, we extract feature maps from the last layer which has a name of conv\textit{j}\_x (\eg, conv4\_6 of Res2Net50 for \texttt{Stage-4}).

For Swin Transformer~\cite{liu2021swin}, it is slightly different from conventional CNN-based models in the fact that they divide an image into tokens.
Unlike ResNet (or Res2Net), there is a layer which works similar to the stem layer (conv\_1), 
a patch partition layer which generates a size of 4 $\times$ 4 patches and aggregates to its spatial dimension for each embedding.
The feature map for the \texttt{Stage-1} is extracted from the patch partition layer, and for \texttt{Stage-j}, where $j > 1$, we extract feature maps from the last layer of stage $j$\footnote{Note that the term `stage $j$' is from~\cite{liu2021swin}.}.
Since Vision Transformers interpret an image as a sequence of patches, we rearrange patches to a 2-dimensional feature map for each stage.

Overall, the hierarchical structure is identical to each other, so we can easily adopt Swin Transformer to other methods, so as we mention in the main paper, we implement 5 SotA models, Chen \etal~\cite{chen2018reverse}, F$^3$Net~\cite{wei2020f3net}, LDF~\cite{wei2020label}, MINet~\cite{pang2020multi} and PA-KRN~\cite{xu2021locate}, and conduct experiments with Res2Net and Swin Transformer backbones.

\begin{figure*}
    \centering
    \includegraphics[width=\textwidth]{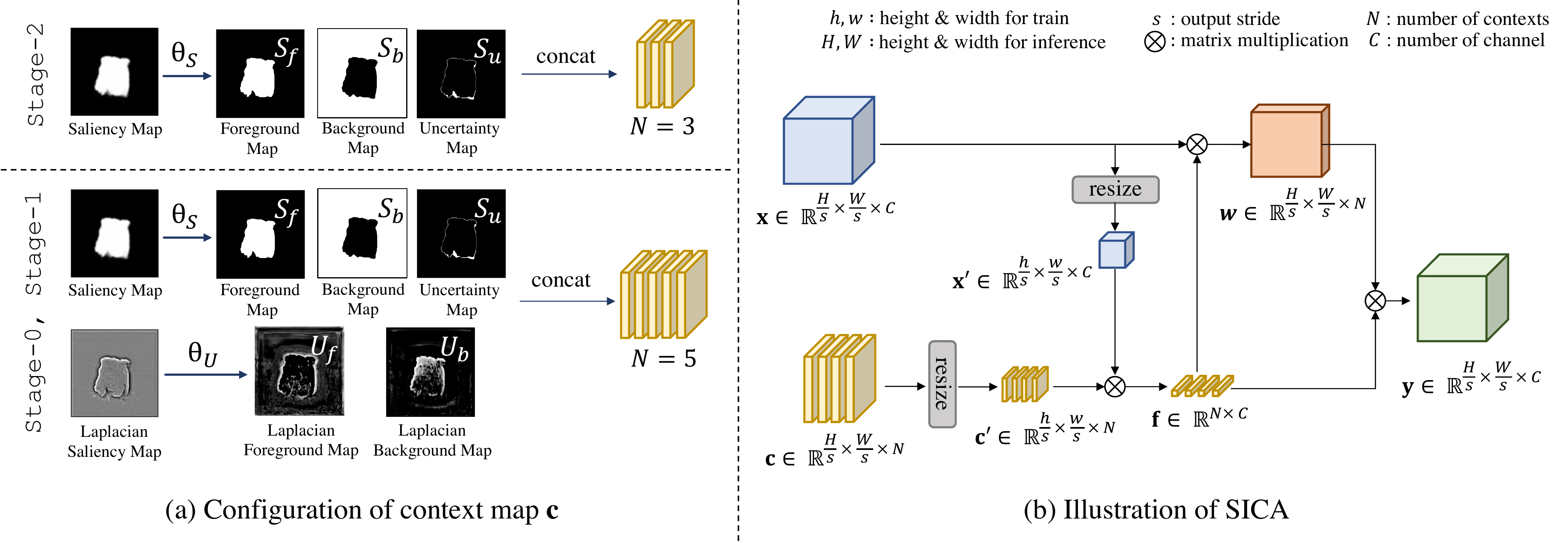} 
    \caption{Context map configuration (a) and details of SICA (b).}
    \label{fig:a1}
\end{figure*}

\subsection{Scale Invariant Context Attention}
In this section, we provide more details of SICA. Since the overall computation of SICA is similar to the OCRNet, we have to obtain soft object regions (or, context maps), which is a coarse segmentation maps corresponds to each class.
However, salient object detection is a class-agnostic segmentation task, so we only have a salient region, and its inverse area. 
UACANet~\cite{kim2021uacanet} first proposed uncertainty area to provide additional context maps, which are often related to the object boundary region.
We notice that the saliency map has an object contextual information, but the Laplacian saliency map also has a boundary context. 
Therefore, we design SICA with additional context clues from the Laplacian saliency map.

However, while UACA set the threshold value as 0.5, we argue that a fixed threshold does not guarantee the optimal saliency map. 
F-measure based loss (FLoss)~\cite{zhao2019optimizing} claims that the optimal threshold obtained with an exhaustive search on the test dataset is impractical for real-world application. 
While they still used exhaustive search after applying FLoss to find the optimal threshold, it is still clear that 0.5 is not the optimal threshold for most cases. 
We also claim that the threshold needs to be adaptive for extracting context information, 
and needs to be different for each stage because the saliency prediction from each stage may have different statistics due to the scale differences. 
So, we choose to learn threshold values for each stage and use them to extract meaningful context regions. 
We illustrate the context map configuration details in~\cref{fig:a1}a.

First, we compute context maps as follows,   
\begin{equation}
    \label{eq:1}
    \begin{gathered}
        S_f =\max(S - \theta_S, 0), \, 
        S_b =\max(\theta_S - S, 0), \\
        S_u = \theta_S - \text{abs}(S - \theta_S),
    \end{gathered}
\end{equation}
\begin{equation}
    \label{eq:2}
    U_f=\max(U - \theta_U, 0), \, U_b=\max(\theta_U - U, 0),
\end{equation}
where we denote trainable threshold $\theta_S$ and $\theta_U$ for input saliency map $S$ and Laplacian saliency map $U$, respectively, and initialize them with 0.5.
$S_f$, $S_b$, and $S_u$ are foreground, background, and uncertainty context maps from $S$. 
Likewise, $U_f$ and $U_b$ are foreground and background context map from $U$. 
Note that because there is no Laplacian saliency map in \texttt{Stage-3}, there are no $U_f$ and $U_b$ for SICA in \texttt{Stage-2}. 
Then, we aggregate context maps for simplicity as follows,
\begin{equation}
    \label{eq:3}
    \textbf{c} = \begin{cases}
            [S_f, S_b, S_u], & \text{if \texttt{Stage-2}}.\\
            [S_f, S_b, S_u, U_f, U_b], & \text{otherwise}.
        \end{cases}
\end{equation}
$\textbf{c}$ consists of three or five context maps depending on its stage (see \cref{fig:a1}a). 

Then, the input feature map from the encoder and the saliency map from the decoder are resized to the size from the training session. 
We denote the size of the input image as $H \times W$, the output stride of the current stage as $s$, and the number of channels and context maps as $C$ and $N$ respectively.
So, the input feature map $\text{x} \in \mathbb{R}^{\frac{H}{s} \times \frac{W}{s} \times C}$ and input context maps $\text{c} \in \mathbb{R}^{\frac{H}{s} \times \frac{W}{s} \times N}$ are resized according to the shape from training time $h \times w$ with bi-linear interpolation.
The resized feature map $\text{x}' \in \mathbb{R}^{\frac{h}{s} \times \frac{w}{s} \times C}$ and context map $\text{c'} \in \mathbb{R}^{\frac{h}{s} \times \frac{w}{s} \times N}$ is used to compute object region representation $\text{f} \in \mathbb{R}^{N \times C}$. 
\begin{equation}
    \label{eq:4}
    \textbf{f}_k = \sum\limits_{(x,y)\in \mathcal{I}} \textbf{c}_k(x,y)\textbf{x}(x,y),
\end{equation}
where $\mathcal{I}$ is a lattice domain of $S$ and $U$.
Since the matrix multiplication is done on the spatial dimension, $\text{f}$ has a same shape with or without resize, yet has better representation ability.

Subsequently, we compute the attention score $w$ by computing the similarity score between $\textbf{f}$ and $\textbf{x}(x,y)$,
\begin{equation}
    \label{eq:5}
    w_k(x,y)=\frac{\exp(\mathcal{T}_\textbf{x}(\textbf{x}(x, y))^\top\mathcal{T}_\textbf{f}(\textbf{f}_k))}{\sum_{l=1}^K \exp(\mathcal{T}_\textbf{x}(\textbf{x}(x, y))^\top\mathcal{T}_\textbf{f}(\textbf{f}_l))},
\end{equation}
where $\mathcal{T}_\textbf{x}(\cdot)$ and $\mathcal{T}_{\textbf{f}}(\cdot)$ denotes transformation functions implemented by consecutive convolution layer, batch normalization, and ReLU activation. 
Lastly, with context representation vector $\textbf{f}$ and attention map $w$ as a weighting factor, we compute a context enhanced feature map $\textbf{y}$ as follows,
\begin{equation}
    \label{eq:6}
    \textbf{y}(x,y)=\mathcal{T}_\textbf{y}(\sum\limits_{l=1}^{K}w_l(x,y)\mathcal{T}_{\textbf{f}'}(\textbf{f}_l)),
\end{equation}
where $\mathcal{T}_\textbf{y}(\cdot)$ and $\mathcal{T}_{\textbf{f}'}(\cdot)$ are transformation functions and $K = 3$ if \texttt{Stage-2}, otherwise, $K = 5$ (see~\cref{eq:3} and~\cref{fig:a1}a). 
The input and output feature maps of SICA are concatenated and forwarded to a simple decoder with convolution blocks to predict the Laplacian saliency map (\cref{fig:a1}b).

\begin{algorithm}[tb]
    \caption{PyTorch pseudocode of image pyramid operations.}
    \label{algo:1}
    \definecolor{codeblue}{rgb}{0.25,0.5,0.5}
     \lstset{
       basicstyle=\fontsize{7.2pt}{7.2pt}\ttfamily\bfseries,
       commentstyle=\fontsize{7.2pt}{7.2pt}\color{codeblue},
       keywordstyle=\fontsize{7.2pt}{7.2pt},
       }
       \begin{lstlisting}[language=python]
        # Module for image pyramid operations (EXPAND, REDUCE)
        # ksize: kernel size for Gaussian filter
        # sigma: standard deviation for Gaussian filter
        # channels: number of channels for pyramid operation 
        #           (For saliency map, set 1. For RGB image, set 3.)
        class ImagePyramid: 
            def __init__(self, ksize=7, sigma=1, channels=1):
                self.ksize = ksize
                self.sigma = sigma
                self.channels = channels
                
                k = cv2.getGaussianKernel(ksize, sigma)
                k = np.outer(k, k)
                k = torch.tensor(k).float()
                self.kernel = k.repeat(channels, 1, 1, 1)
            
            # call to use GPU
            def cuda(self):
                self.kernel = self.kernel.cuda()
                return self
                
            # EXPAND operation
            def expand(self, x):
                z = torch.zeros_like(x)
                x = torch.cat([x, z, z, z], dim=1)
                x = F.pixel_shuffle(x, 2)
                x = F.pad(x, (self.ksize // 2, ) * 4, mode='reflect')
                x = F.conv2d(x, self.kernel * 4, groups=self.channels)
                return x
            
            # REDUCE operation
            def reduce(self, x):
                x = F.pad(x, (self.ksize // 2, ) * 4, mode='reflect')
                x = F.conv2d(x, self.kernel, groups=self.channels)
                return x[:, :, ::2, ::2]
    \end{lstlisting}
\end{algorithm}

\subsection{Implementation of Image Pyramid Operations.}
    We implement EXPAND and REDUCE operations~\cite{burt1983laplacian} in Pytorch~\cite{paszke2019pytorch}. We provide a source code of both operations in~\cref{algo:1}.
    We obtain the 1D Gaussian kernel with \texttt{cv2.getGaussianKernel} from OpenCV~\cite{opencv_library}, and use outer operation to generate the 2D Gaussian kernel.
    For EXPAND operation, we use pixel-shuffle operation from PyTorch, and we used even indices for REDUCE operation.

    For pyramid blending, we use \texttt{cv2.getStructuringElement} function with \texttt{cv2.MORPH\_ELIPSE} argument for dilation and erosion.
    Also, we use \texttt{dilation} and \texttt{erosion} functions from \texttt{kornia.morphology}~\cite{riba2020kornia}.
    The kernel size for dilation and erosion is set to 5, 9, and 17 for \texttt{Stage-2}, \texttt{Stage-1}, and \texttt{Stage-0} respectively.

\section{Experiments}

\subsection{Ablation Studies}
\noindent
\textbf{Gaussian Filter in Image Pyramid.} Even the kernel size $k$ and the standard deviation $\sigma$ of the Gaussian kernel $g$ for image pyramid operation is usually set to 5 and 1 respectively, 
we compare our image pyramid operations with different kernel sizes and standard deviations. 
Results in~\cref{tab:a1} shows that when the $\sigma$ gets larger, the overall performance decreases. 
Furthermore, when $k$ is 7 and $\sigma$ is set to 1, we obtain the best result among different settings.

\begin{table}
    \setlength{\tabcolsep}{2.5pt}
    \centering
    \caption{Left: Ablation study of kernel size $k$ and the standard deviation $\sigma$ of the Gaussian kernel $G(k, \sigma)$ for image pyramid operations of InSPyReNet (Res2Net50) on DUTS-TE and DUT-OMRON\@.
    Right: Ablation study of training strategies for InSPyReNet (SwinB) on DAVIS-S and HRSOD-TE\@.
    \textit{pred} and \textit{gt} denotes the image pyramid structure applied in prediction and ground truth respectively. 
    S.G. denotes Stop-Gradient. 
    $\mathcal{L}_{pc}$ denotes pyramidal consistency loss.}
    \resizebox{\linewidth}{!}{
        \begin{tabular}{cc|ccc|ccc}
            \hline \hline
            \multirow{2}{*}{$k$}
            & \multirow{2}{*}{$\sigma$}
            & \multicolumn{3}{c|}{DUTS-TE}             
            & \multicolumn{3}{c}{DUT-OMRON} \\
            & & $S_\alpha$ & $F_{\max}$ & MAE & $S_\alpha$ & $F_{\max}$ & MAE \\ \hline
            5 & 1 & 0.902 & 0.890 & 0.037 & 0.839 & 0.783 & 0.059 \\
            5 & 3 & 0.897 & 0.882 & 0.041 & 0.837 & 0.779 & 0.059 \\
            5 & 5 & 0.888 & 0.876 & 0.042 & 0.834 & 0.770 & 0.060 \\
            \hline
            \rowgray
            7 & 1 & \textbf{0.904} & \textbf{0.892} & \textbf{0.035} & \textbf{0.845} & \textbf{0.791} & 0.059 \\
            7 & 3 & 0.897 & 0.884 & 0.037 & 0.841 & 0.777 & \textbf{0.058} \\
            7 & 5 & 0.888 & 0.878 & 0.038 & 0.833 & 0.774 & 0.061 \\
            \hline \hline
        \end{tabular}
        
        \hspace{1em}

        \begin{tabular}{cccc|cccc|cccc}
            \hline \hline
            \multirow{2}{*}{\textit{pred}}
            & \multirow{2}{*}{\textit{gt}}
            & \multirow{2}{*}{S.G.}
            & \multirow{2}{*}{$\mathcal{L}_{pc}$}
            & \multicolumn{4}{c|}{DAVIS-S}             
            & \multicolumn{4}{c}{HRSOD-TE} \\
            & & & & $S_\alpha\uparrow$ & $F_{\max}\uparrow$ & MAE$\downarrow$ & mBA$\uparrow$ & $S_\alpha\uparrow$ & $F_{\max}\uparrow$ & MAE$\downarrow$ & mBA$\uparrow$ \\ \hline
                       &            &            &            & 0.935 & 0.937 & 0.016 & 0.693 & 0.931 & 0.933 & 0.023 & 0.682 \\
            \checkmark &            &            &            & 0.937 & 0.939 & 0.014 & 0.714 & 0.934 & 0.937 & 0.019 & 0.712 \\
                       & \checkmark &            &            & 0.938 & 0.935 & 0.016 & 0.699 & 0.932 & 0.931 & 0.022 & 0.695 \\
            \checkmark & \checkmark &            &            & 0.945 & 0.942 & 0.014 & 0.719 & 0.944 & 0.942 & 0.017 & 0.715 \\
            \checkmark & \checkmark & \checkmark &            & 0.955 & \textbf{0.959} & 0.010 & 0.727 & 0.947 & 0.948 & 0.018 & 0.729 \\
            \rowgray
            \checkmark & \checkmark & \checkmark & \checkmark & \textbf{0.962} & 0.959 & \textbf{0.009} & \textbf{0.743} & \textbf{0.952} & \textbf{0.949} & \textbf{0.016} & \textbf{0.738} \\
            \hline \hline
        \end{tabular}

        }
    \label{tab:a1}
  \end{table}

\noindent
\textbf{Training Strategies.} To demonstrate the effect of our training strategies such as supervision under image pyramid structure (\textit{pred} and \textit{gt}), pyramidal consistency loss ($\mathcal{L}_{pc}$), stop-gradient(S.G.), 
we train InSPyReNet with different settings and evaluate on HR benchmarks. 
Results in~\cref{tab:a1} shows that without image pyramid structure on prediction branch (\textit{pred}) shows unsatisfactory results in terms of mBA.
This is because while other strategies (\textit{gt}, S.G., $\mathcal{L}_{pc}$) are considered as supervision strategies, \textit{pred} is embedded as a model architecture.
With other strategies, we can notice that S.G. provides some improvements in terms of SOD metrics, which means it gives more stable results for pyramid blending.
Also, $\mathcal{L}_{pc}$ shows extra improvements for mBA, which means it ensures the image pyramid structure in a sense of high-frequency residual information of Laplacian image.

\begin{figure*}
    \centering
    \includegraphics[width=\textwidth]{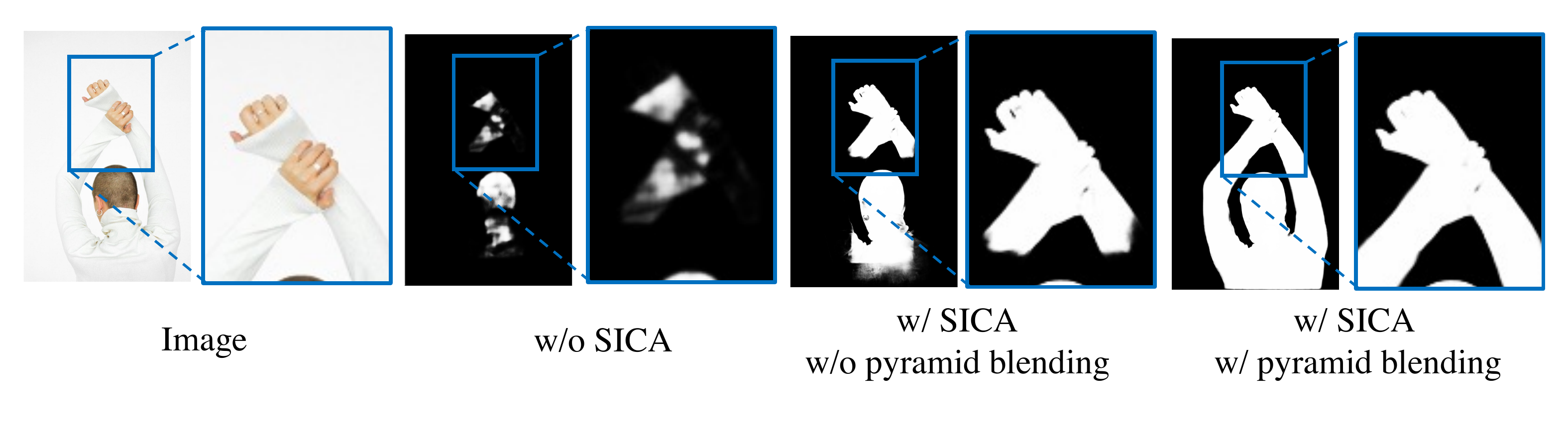} 
    \caption{Visual demonstration for ablation study of SICA. The sample is taken from AIM-500~\cite{Li2021AIM}.}
    \label{fig:a2}
\end{figure*}

\noindent
\textbf{SICA.}
As shown in~\cref{fig:a2}, methods without SICA generates unpleasant artifacts on the boundary areas, which can be interpreted as the failure of Laplacian images of the saliency map from SICA.
On the other hand, InSPyReNet with SICA shows detailed predictions, but without pyramid blending, it misses some major object parts. 
With SICA and pyramid blending shows best results, showing less failure in terms of both capturing the whole salient object body parts and high-frequency object boundary details.

\begin{table}
    \setlength{\tabcolsep}{2.5pt}
    \centering
    \caption{Quantitative results of applying pyramid blending for previous pyramid-based SOD method (Chen \etal~\cite{chen2018reverse}) on three HR and two LR benchmarks. P.B. denotes pyramid blending.
    $\uparrow$ indicates larger the better, and $\downarrow$ indicates smaller the better.}
    \resizebox{\linewidth}{!}{
        \begin{tabular}{c|cccc|cccc|cccc}
            \hline \hline
            \multirow{2}{*}{Algorithms} & \multicolumn{4}{c|}{DAVIS-S} & \multicolumn{4}{c|}{HRSOD-TE} & \multicolumn{4}{c}{UHRSD-TE} \\ 
            & $S_\alpha\uparrow$ & $F_{\max}\uparrow$ & MAE$\downarrow$ & mBA$\uparrow$
            & $S_\alpha\uparrow$ & $F_{\max}\uparrow$ & MAE$\downarrow$ & mBA$\uparrow$ 
            & $S_\alpha\uparrow$ & $F_{\max}\uparrow$ & MAE$\downarrow$ & mBA$\uparrow$ \\ \hline\hline
            \multicolumn{13}{c}{w/o pyramid blending} \\ \hline\hline
            Chen \etal~\cite{chen2018reverse} & 0.934 & 0.925 & 0.018 & 0.697 & 0.915 & 0.907 & 0.032 & 0.684 & 0.915 & 0.919 & 0.034 & 0.712 \\
            \textit{Ours}              & 0.953 & 0.949 & 0.013 & 0.705 & 0.945 & 0.941 & 0.019 & 0.700 & 0.927 & 0.932 & 0.032 & 0.724 \\ \hline\hline
            \multicolumn{13}{c}{w/ pyramid blending} \\ \hline\hline
            Chen \etal~\cite{chen2018reverse} & 0.916 & 0.893 & 0.023 & 0.583 (\textcolor{red}{-16.4\%})  & 0.909 & 0.891 & 0.033 & 0.590 (\textcolor{red}{-13.7\%})  & 0.903 & 0.906 & 0.042 & 0.590 (\textcolor{red}{-17.1\%})\\ 
            \textit{Ours}              & 0.962 & 0.958 & 0.009 & 0.732 (\textcolor{green}{+3.8\%}) & 0.952 & 0.955 & 0.015 & 0.732 (\textcolor{green}{+4.6\%}) & 0.932 & 0.938 & 0.029 & 0.741 (\textcolor{green}{+2.3\%})\\ 
            \hline \hline
        \end{tabular}}
    \label{tab:a2}
\end{table}

\begin{figure*}
    \centering
    \includegraphics[width=\textwidth]{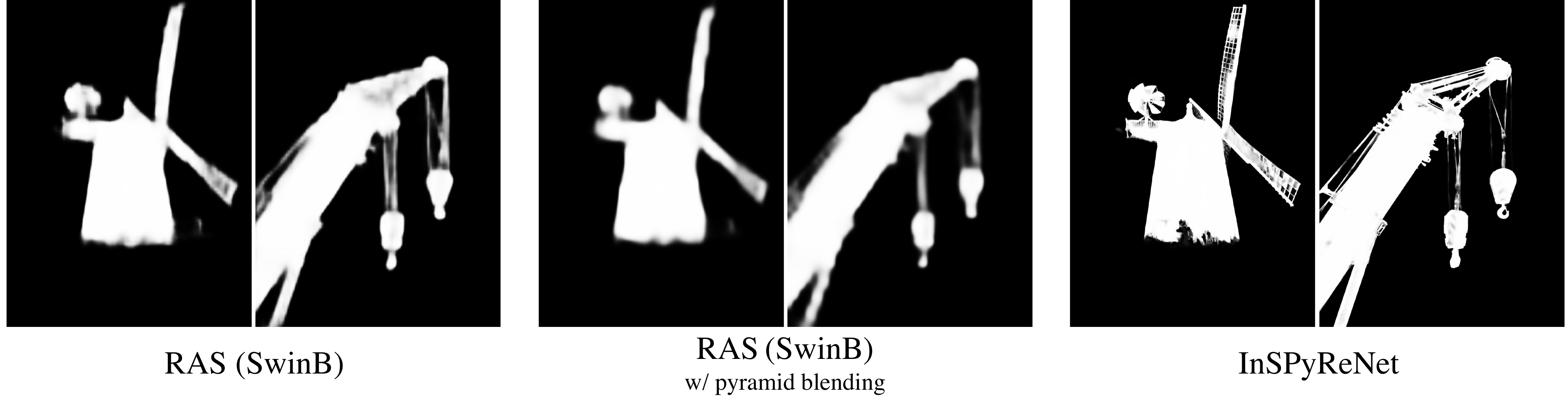} 
    \caption{Visual comparison of applying pyramid blending for previous pyramid-based SOD method (Chen \etal) and InSPyReNet. Samples are taken from UHRSD~\cite{xie2022pyramid}. \textit{Best viewed by zooming in.}}
    \label{fig:a3}
\end{figure*}

\subsection{Applying pyramid blending to previous Image Pyramid based Model}
It is easy to apply pyramid blending if the base model outputs image pyramid of saliency map same as InSPyReNet.
We choose Chen \etal~\cite{chen2018reverse} since it has a great reproducibility and has a pyramid structure.
We trained Chen \etal with same backbone (SwinB) for fair comparison.
As shown in~\cref{tab:a2}, pyramid blending for Chen \etal worsen results especially for the mBA measure which is a major reason for pyramid blending.
We also provide some qualitative results of Chen \etal with pyramid blending in~\cref{fig:a3}, which shows some clear degradation.
Thus, without a well-defined image pyramid based model designed for image blending, it cannot be used for HR prediction.

\subsection{Training InSPyReNet with HR datasets}
To demonstrate the potential of our method, we utilize HR datasets (HRSOD-TR, UHRSD-TR) alongside DUTS-TR for our training dataset. 
First, we use HR datasets for training with fixed LR scale (\ie, $384 \times 384$), meaning that we do not use HR annotations and regard them as another LR datasets.
Results show that our method well generalizes to the HR prediction with extra LR datasets (\cref{tab:a3}).
Note that we do not include this experiments for our final results since even we resize HR datasets into LR scale, 
annotations are still remains high-quality thanks to the interpolation method, so it is not fair to claim that we are using only LR datasets.

\begin{figure*}
    \centering
    \includegraphics[width=\textwidth]{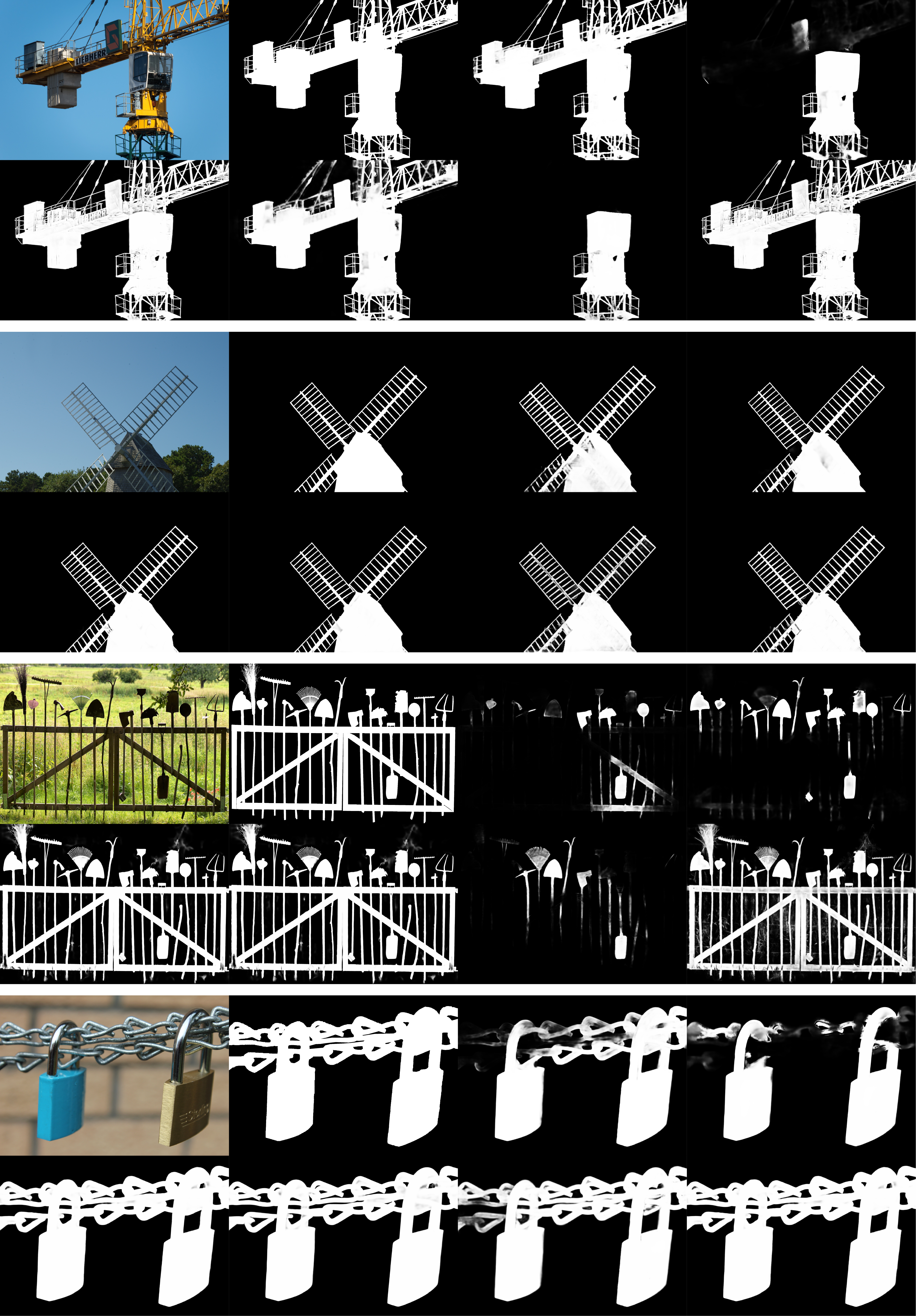} 
    \caption{Visual comparison of PGNet~\cite{xie2022pyramid} (H, U) and our methods trained with HR datasets. Results are ordered as image, ground truth, PGNet, and \textit{Ours}(D, H$^*$) in the first row, and \textit{Ours}(H$^*$, U$^*$), \textit{Ours}(D, H$^*$, U$^*$), \textit{Ours}(D, H), \textit{Ours}(H, U) in the second row from left to right for each sample. \textit{Best viewed by zooming in.}}
    \label{fig:a4}
\end{figure*}
\begin{table}
    \setlength{\tabcolsep}{2.5pt}
    \centering
    \caption{Quantitative results on three HR and two LR benchmarks for training with extra HR datasets. 
    D: DUTS-TR, H: HRSOD-TR, U: UHRSD-TR. 
    The best results for each metric are denoted as \textbf{bold}. 
    $\uparrow$ indicates larger the better, and $\downarrow$ indicates smaller the better.
    $^*$ indicates that the dataset is resized into LR scale.}
    \resizebox{\linewidth}{!}{
        \begin{tabular}{c|cccc|cccc|cccc|ccc|ccc}
            \hline \hline
            \multirow{3}{*}{\shortstack{Train\\Datasets}} & \multicolumn{12}{c|}{HR benchmarks} & \multicolumn{6}{c}{LR benchmarks} \\ \cline{2-19}
            & \multicolumn{4}{c|}{DAVIS-S} & \multicolumn{4}{c|}{HRSOD-TE} & \multicolumn{4}{c|}{UHRSD-TE} & \multicolumn{3}{c|}{DUTS-TE} & \multicolumn{3}{c}{DUT-OMRON}  \\
            & $S_\alpha\uparrow$ & $F_{\max}\uparrow$ & MAE$\downarrow$ & mBA$\uparrow$ & $S_\alpha\uparrow$ & $F_{\max}\uparrow$ & MAE$\downarrow$ & mBA$\uparrow$ & $S_\alpha\uparrow$ & $F_{\max}\uparrow$ & MAE$\downarrow$ & mBA$\uparrow$ & $S_\alpha\uparrow$ & $F_{\max}\uparrow$ & MAE$\downarrow$ & $S_\alpha\uparrow$ & $F_{\max}\uparrow$ & MAE$\downarrow$ \\ \hline\hline
            \multicolumn{19}{c}{PGNet~\cite{xie2022pyramid}} \\ \hline \hline
            H,U           & 0.954 & 0.956 & 0.010 & 0.730 & 0.938 & 0.939 & 0.020 & 0.727 & 0.935 & 0.930 & 0.026 & 0.765 & 0.861 & 0.828 & 0.038 & 0.790 & 0.727 & 0.059 \\ \hline \hline
            \multicolumn{19}{c}{\textit{Ours} Trained with LR scale (\ie, $384 \times 384$)} \\ \hline \hline
            D, H$^*$      & 0.963 & 0.966 & 0.008 & 0.744 & 0.958 & 0.958 & 0.014 & 0.752 & 0.937 & 0.945 & 0.027 & 0.754 & \textbf{0.936} & \textbf{0.934} & \textbf{0.022} & 0.878 & 0.836 & 0.044 \\
            H$^*$,U$^*$   & 0.963 & 0.967 & 0.008 & 0.732 & 0.947 & 0.945 & 0.020 & 0.741 & 0.949 & 0.956 & \textbf{0.020} & 0.765 & 0.925 & 0.922 & 0.028 & 0.874 & 0.835 & 0.048 \\
            D,H$^*$,U$^*$ & 0.970 & 0.972 & \textbf{0.007} & 0.743 & 0.951 & 0.951 & 0.018 & 0.748 & 0.950 & \textbf{0.957} & \textbf{0.020} & 0.767 & 0.931 & 0.928 & 0.024 & \textbf{0.880} & \textbf{0.837} & \textbf{0.042} \\ \hline \hline
            \multicolumn{19}{c}{\textit{Ours} Trained with HR scale (\ie, $1024 \times 1024$)} \\ \hline \hline
            D,H           & 0.972 & 0.976 & \textbf{0.007} & \textbf{0.770} & \textbf{0.960} & \textbf{0.957} & \textbf{0.014} & 0.766 & 0.936 & 0.938 & 0.028 & 0.785 & 0.934 & 0.927 & 0.023 & 0.859 & 0.799 & 0.049 \\
            H,U           & \textbf{0.973} & \textbf{0.977} & \textbf{0.007} & \textbf{0.770} & 0.956 & 0.956 & 0.018 & \textbf{0.771} & \textbf{0.953} & \textbf{0.957} & \textbf{0.020} & \textbf{0.812} & \textbf{0.936} & 0.932 & 0.024 & 0.872 & 0.823 & 0.046 \\
            \hline \hline
        \end{tabular}}
    \label{tab:a3}
\end{table}

Moreover, to understand the potential of InSPyReNet as is, we trained our method with HR datasets in HR scale.
In this case, we do not deploy pyramid blending since we are not training in LR scale, and hence there is no LR pyramid to merge with.
For HR training, we follow the training size from~\cite{xie2022pyramid} (\ie, $1024 \times 1024$).
As shown in~\cref{tab:a3}, our method shows great results with fully supervised manner for HR prediction, even we do not specifically design InSPyReNet for HR prediction without pyramid blending.
This experiment shows that with simple image pyramid structure from our method can further be utilized for HR prediction with HR datasets.
Overall, results trained with HR scale shows better performance especially for mBA, but slightly worse on LR benchmarks than results trained on LR scale.
From this experiment, although the performance on LR and HR benchmarks tends to prefer the corresponding training scale, InSPyReNet well adopts to each other.

We also provide qualitative results in~\cref{fig:a4}. 
Compared to the same setting of PGNet~\cite{xie2022pyramid} trained with HRSOD-TR and UHRSD-TR, our method substantially outperforms the quality of high-frequency details of saliency maps.
Moreover, we can notice that without UHRSD-TR whether we train with LR or HR scale, we cannot expect good results for complex scenes (first and third samples in~\cref{fig:a4}).
This is because while DUTS-TR and HRSOD-TR are in favor of ``centered'' objects (\eg, our methods without UHRSD-TR in the first sample), while UHRSD-TR more focus on complex details which usually cover the whole image like thrid sample in~\cref{fig:a4}.

\begin{table}
    \setlength{\tabcolsep}{2.5pt}
    \centering
    \caption{Quantitative results of IS-Net and our method trained with DIS5K. We trained our model with LR scale (\eg, $384 \times 384$) and HR scale (\eg, $1024 \times 1024$).}
    \resizebox{\linewidth}{!}{
        \begin{tabular}{cccc|cccc|cccc|cccc|cccc}
            \hline \hline
            \multicolumn{4}{c|}{DIS-VD} & \multicolumn{4}{c|}{DIS-TE1} & \multicolumn{4}{c|}{DIS-TE2} & \multicolumn{4}{c|}{DIS-TE3} & \multicolumn{4}{c}{DIS-TE4} \\
            $S_\alpha\uparrow$ & $F_{\max}\uparrow$ & $HCE_{\gamma}\downarrow$ & mBA$\uparrow$ & $S_\alpha\uparrow$ & $F_{\max}\uparrow$ & $HCE_{\gamma}\downarrow$ & mBA$\uparrow$ & $S_\alpha\uparrow$ & $F_{\max}\uparrow$ & $HCE_{\gamma}\downarrow$ & mBA$\uparrow$ & $S_\alpha\uparrow$ & $F_{\max}\uparrow$ & $HCE_{\gamma}\downarrow$ & mBA$\uparrow$ & $S_\alpha\uparrow$ & $F_{\max}\uparrow$ & $HCE_{\gamma}\downarrow$ & mBA$\uparrow$ \\ \hline\hline 
            \multicolumn{20}{c}{IS-Net~\cite{qin2022highly}} \\ \hline \hline
            0.813 & 0.791 & 1116 & 0.741 & 0.787 & 0.740 & 149 & 0.736 & 0.823 & 0.799 & 340 & 0.740 & 0.836 & 0.830 & 687 & 0.746 & 0.830 & 0.827 & 2888 & 0.743 \\ \hline\hline
            \multicolumn{20}{c}{\textit{Ours} Trained with LR scale (\ie, $384 \times 384$)} \\ \hline \hline
            0.887 & 0.876 & 905 & 0.765 & 0.862 & 0.834 & 148 & 0.745 & 0.893 & 0.881 & 316 & 0.759 & 0.902 & 0.904 & 582 & 0.774 & 0.891 & 0.892 & \textbf{2243} & 0.779  \\ \hline\hline
            \multicolumn{20}{c}{\textit{Ours} Trained with HR scale (\ie, $1024 \times 1024$)} \\ \hline \hline
            \textbf{0.900} & \textbf{0.889} & \textbf{904} & \textbf{0.800} & \textbf{0.873} & \textbf{0.845} & \textbf{110} & \textbf{0.797} & \textbf{0.905} & \textbf{0.894} & \textbf{255} & \textbf{0.803} & \textbf{0.918} & \textbf{0.919} & \textbf{522} & \textbf{0.808} & \textbf{0.905} & \textbf{0.905} & 2336 & \textbf{0.799} \\ \hline\hline
        \end{tabular}}
    \label{tab:a5}
\end{table}

\begin{figure*}
    \centering
    \includegraphics[width=\textwidth]{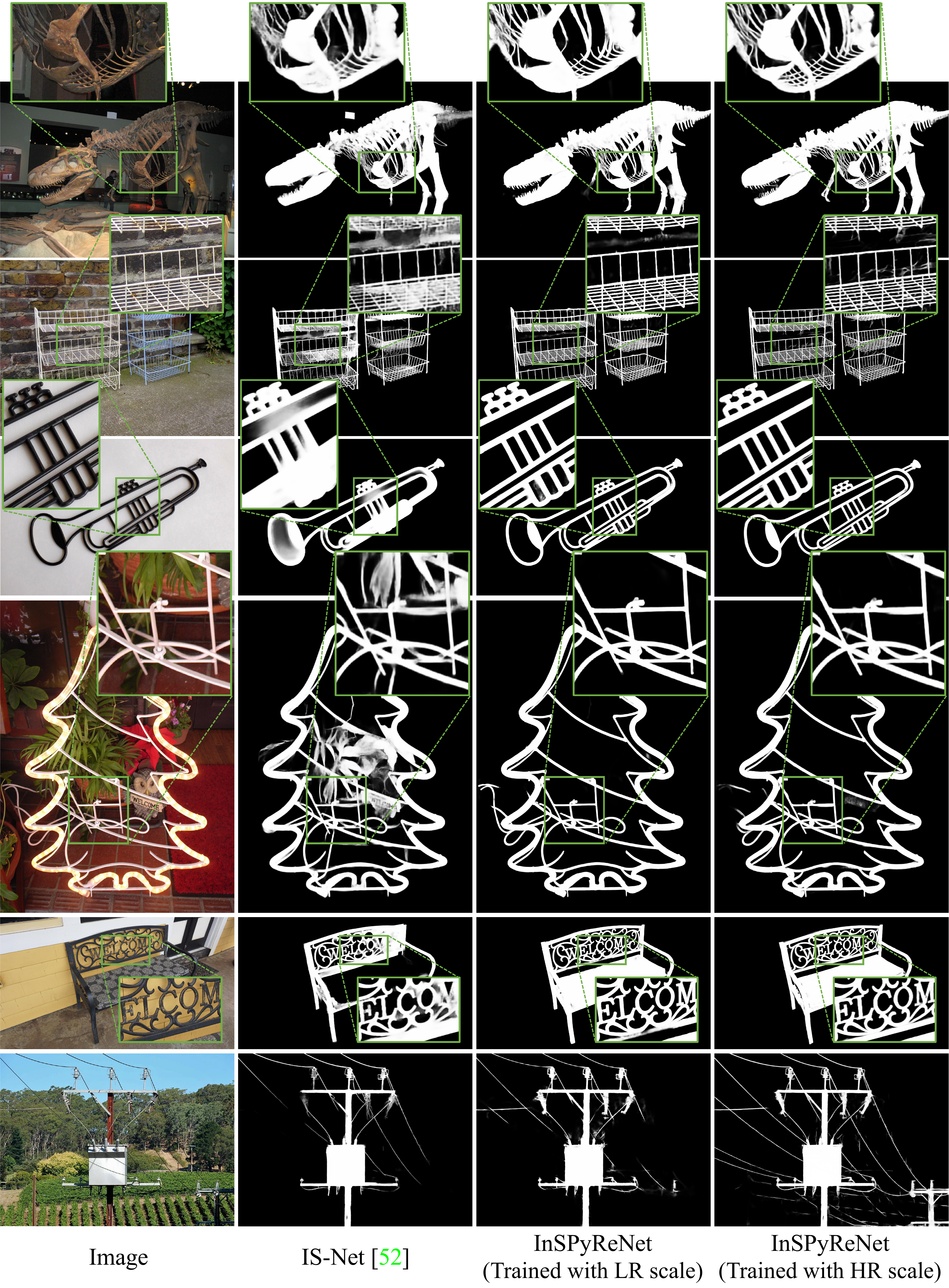} 
    \caption{Visual comparison of IS-Net~\cite{qin2022highly} and our method trained with DIS5K. We trained our model with both LR and HR scales each. \textit{Best viewed by zooming in.}}
    \label{fig:a6}
\end{figure*}

\subsection{Dichotomous Image Segmentation (DIS5K)}

We trained our model and compare to the baseline model of DIS5K, IS-Net~\cite{qin2022highly} which is currently shows SotA performance.
To demonstrate the potential of our method which can be trained with LR scale (\eg, $384 \times 384$) to boost up training time and reduce required resources, we trained our model with both LR and HR scales.
In~\cref{tab:a5}, our method shows regardless of training scale shows great performance compared to the baseline model, IS-Net~\cite{qin2022highly}.
Moreover, in~\cref{fig:a6}, we provide qualitative results to demonstrate that our model can produce detailed output regardless of the training scale.

\section{Discussion}
\begin{table}
    \setlength{\tabcolsep}{2.5pt}
    \centering
    \caption{Comparison of boundary quality measures (BDE~\cite{freixenet2002yet}, mBA~\cite{cheng2020cascadepsp}, BIoU~\cite{cheng2021boundary}) with HR SOD methods. D: DUTS-TR, H: HRSOD-TR, U: UHRSD-TR. 
    Three best results in order except our method are colored as \textcolor{red}{red}, \textcolor{blue}{blue}, and \textcolor{green}{green}.}
        \begin{tabular}{c|c|ccc|ccc}
            \hline \hline
            \multirow{2}{*}{Algorithms} & \multirow{2}{*}{\shortstack{Train \\ Datasets}}
            & \multicolumn{3}{c|}{DAVIS-S} & \multicolumn{3}{c}{HRSOD-TE}\\
            & & BDE & mBA & BIoU 
              & BDE & mBA & BIoU \\ \hline \hline

            Zeng \etal~\cite{zeng2019towards}      & D, H & 44.359 & 0.618 & 0.662 & 88.017 & 0.623 & 0.659 \\
            Tang \etal~\cite{tang2021disentangled} & D, H & \textcolor{blue}{14.266} & \textcolor{blue}{0.716} & \textcolor{green}{0.785} & \textcolor{blue}{46.495} & \textcolor{green}{0.693} & 0.744 \\
            PGNet~\cite{xie2022pyramid}            & D    & 34.957 & \textcolor{green}{0.707} & 0.769 & \textcolor{green}{46.923} & \textcolor{green}{0.693} & \textcolor{green}{0.749} \\
            PGNet~\cite{xie2022pyramid}            & D, H & \textcolor{green}{14.463} & \textcolor{blue}{0.716} & \textcolor{blue}{0.790} & \textcolor{red}{45.292} & \textcolor{blue}{0.714} & \textcolor{blue}{0.772} \\
            PGNet~\cite{xie2022pyramid}            & H, U & \textcolor{red}{12.725} & \textcolor{red}{0.730} & \textcolor{red}{0.814} & 57.147 & \textcolor{red}{0.727} & \textcolor{red}{0.781} \\
            \hline \hline
            \textit{Ours}                          & D    &    -   & 0.743 & 0.850 &    -   & 0.738 & 0.826 \\ 
            
            \hline \hline
        \end{tabular}
    \label{tab:a4}
\end{table}

\subsection{Selection of Boundary Metrics}
Although many SOD methods dedicated to the HR benchmarks~\cite{zeng2019towards,tang2021disentangled,xie2022pyramid} use Boundary Displacement Error (BDE)~\cite{freixenet2002yet}, 
we suggest to use mean boundary accuracy (mBA)~\cite{cheng2020cascadepsp} instead for following reasons.
First, it is substantially outdated metric for measuring boundary quality since BDE was proposed in 2002, when image segmentation methods highly depended on low-level signal analysis of the given image.
Now we're in the era of deep learning. We can easily generate more accurate, high-quality segmentation results, and hence need to use metrics like mBA or Boundary IoU (BIoU)~\cite{cheng2021boundary}.
Second, we could not find any of official, non-official implementation related to the BDE, and the only source that we found is unable to access.
While, mBA and BIoU have official implementations. 
We report mBA in the main paper because it does not require modification for SOD since~\cite{cheng2020cascadepsp} also works for a binary segmentation map like SOD, 
while BIoU requires major modification since the official code only provides BIoU embedded in the evaluation codes for Average Precision and Panoptic Quality.
Third, the evaluation results with BDE shows inconsistent while mBA and BIoU shows consistent results as shown in~\cref{tab:a4}.
On the other hand, mBA and BIoU shows consistent results across different HR methods. Thus, we use mBA for boundary metric.

\begin{figure*}
    \centering
    \includegraphics[width=\textwidth]{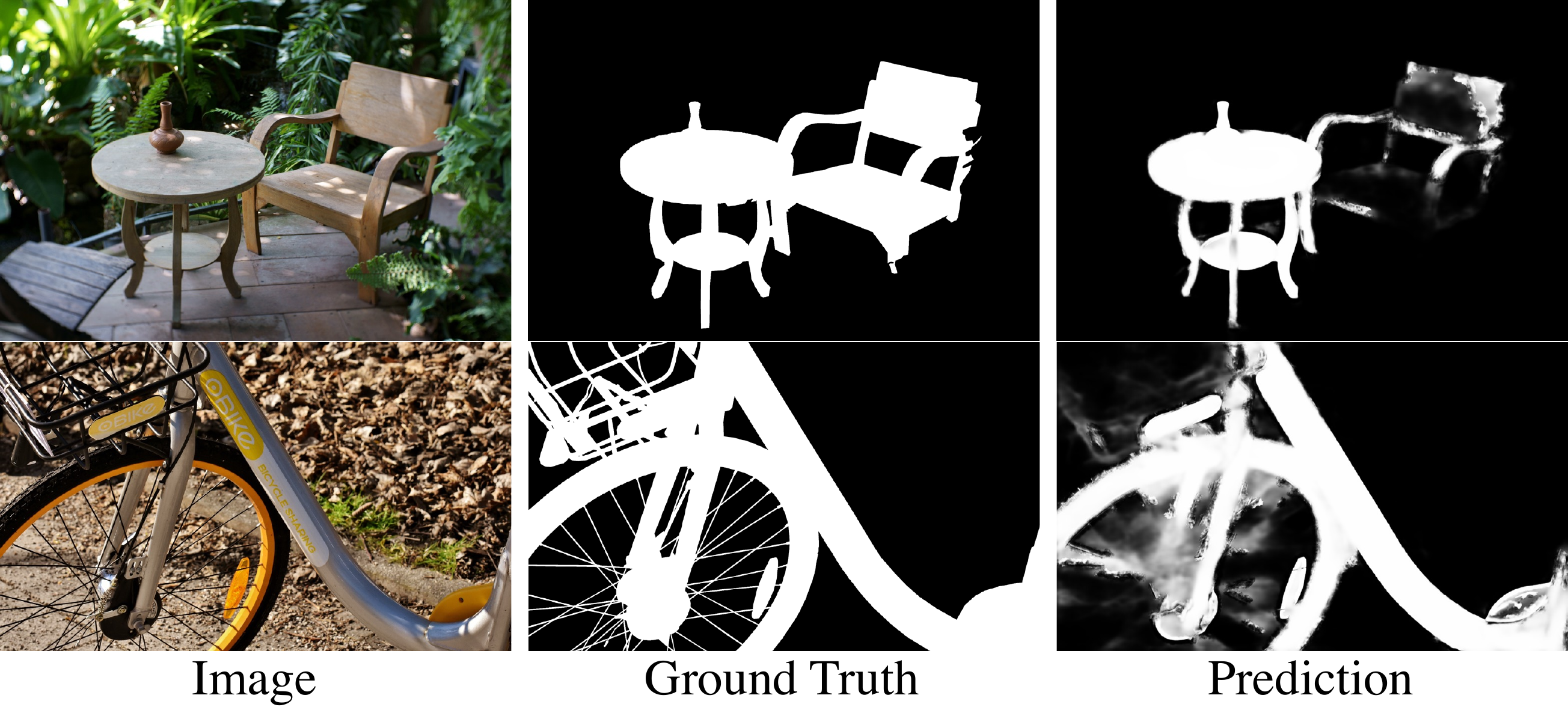} 
    \caption{Visual illustration for the failure case (first row: global context failure, second row: local detail failure) of InSPyReNet. \textit{Best viewed by zooming in.}}
    \label{fig:a5}
\end{figure*} 

\subsection{Potential Vulnerability of InSPyReNet}
While we show that our InSPyReNet can produce high-quality results in HR benchmarks, we can notice that there are some failure cases in terms of two different aspects.
First, as shown in the first row from~\cref{fig:a5}, if the LR branch in pyramid blending fails to predict the saliency object, it suffers from global context failure.
Second, even if the LR branch in pyramid blending successfully predict the saliency branch, we still have a chance to fail reconstructing local details when the HR branch fails to generate high-frequency details.
In the second row from~\cref{fig:a5}, the LR branch detected the body part of the bicycle, but from HR branch, it fails to predict spokes of the wheel and details of the front basket.

\end{document}